\documentclass[sigconf]{acmart}
\usepackage{subfig}
\usepackage{amsmath,amsfonts}
\usepackage{algorithm}
\usepackage{algorithmic}
\usepackage[algo2e]{algorithm2e}
\usepackage{multirow}
\usepackage[export]{adjustbox}
\usepackage{xcolor}
\AtBeginDocument{%
  \providecommand\BibTeX{{%
    \normalfont B\kern-0.5em{\scshape i\kern-0.25em b}\kern-0.8em\TeX}}}

\copyrightyear{2023}
\acmYear{2023}
\setcopyright{acmlicensed}\acmConference[MM '23]{Proceedings of the 31st ACM International Conference on Multimedia}{October 29-November 3, 2023}{Ottawa, ON, Canada}
\acmBooktitle{Proceedings of the 31st ACM International Conference on Multimedia (MM '23), October 29-November 3, 2023, Ottawa, ON, Canada}
\acmPrice{15.00}
\acmDOI{10.1145/3581783.3611730}
\acmISBN{979-8-4007-0108-5/23/10}

\begin{document}
\title{DocDiff: Document Enhancement via Residual Diffusion Models}
% \acmSubmissionID{175}
% \settopmatter{printacmref=false}
% \setcopyright{acmcopyright}
% \copyrightyear{2023}
% \acmYear{2023}
% \acmDOI{XXXXXXX.XXXXXXX}
% %% These commands are for a PROCEEDINGS abstract or paper.
% \acmConference[MM '23]{Proceedings of the 31th ACM International Conference on Multimedia}{October 29, 2023}{Ottawa, Canada}
% \acmBooktitle{Proceedings of the 31th ACM International Conference on Multimedia (MM '23), October 29, 2023, Ottawa, Canada}
% \acmPrice{15.00}
% \acmISBN{978-1-4503-XXXX-X/18/06}
%%
%% The "author" command and its associated commands are used to define
%% the authors and their affiliations.
%% Of note is the shared affiliation of the first two authors, and the
%% "authornote" and "authornotemark" commands
%% used to denote shared contribution to the research.

\author{Zongyuan Yang}
\affiliation{%
  \institution{Beijing University of Posts and Telecommunications}
  \city{Beijing}
  \country{China}
}
\email{yangzongyuan0@bupt.edu.cn}

\author{Baolin Liu}
\affiliation{%
  \institution{Beijing University of Posts and Telecommunications}
  \city{Beijing}
  \country{China}
}
\email{baolin@bupt.edu.cn}

\author{Yongping Xiong}
\authornote{Corresponding author}
\affiliation{%
  \institution{Beijing University of Posts and Telecommunications}
  \city{Beijing}
  \country{China}
}
\email{ypxiong@bupt.edu.cn}

\author{Lan Yi}
\affiliation{%
  \institution{Beijing University of Posts and Telecommunications}
  \city{Beijing}
  \country{China}
}

\author{Guibin Wu}
\affiliation{%
  \institution{Beijing University of Posts and Telecommunications}
  \city{Beijing}
  \country{China}
}

\author{Xiaojun Tang}
\affiliation{%
  \institution{Beijing University of Posts and Telecommunications}
  \city{Beijing}
  \country{China}
}

\author{Ziqi Liu}
\affiliation{%
  \institution{Beijing University of Posts and Telecommunications}
  \city{Beijing}
  \country{China}
}

\author{Junjie Zhou}
\affiliation{%
  \institution{Beijing University of Posts and Telecommunications}
  \city{Beijing}
  \country{China}
}

\author{Xing Zhang}
\affiliation{%
  \institution{Artificial Intelligence Lab of China Resources Digital Technology}
  \city{Guangdong}
  \country{China}
}
%%
%% By default, the full list of authors will be used in the page
%% headers. Often, this list is too long, and will overlap
%% other information printed in the page headers. This command allows
%% the author to define a more concise list
%% of authors' names for this purpose.
\renewcommand{\shortauthors}{Trovato and Tobin, et al.}

%%
%% The abstract is a short summary of the work to be presented in the
%% article.
\begin{abstract}
Removing degradation from document images not only improves their visual quality and readability, but also enhances the performance of numerous automated document analysis and recognition tasks. However, existing regression-based methods optimized for pixel-level distortion reduction tend to suffer from significant loss of high-frequency information, leading to distorted and blurred text edges. To compensate for this major deficiency, we propose DocDiff, the first diffusion-based framework specifically designed for diverse challenging document enhancement problems, including document deblurring, denoising, and removal of watermarks and seals. DocDiff consists of two modules: the \textbf{C}oarse \textbf{P}redictor (CP), which is responsible for recovering the primary low-frequency content, and the \textbf{H}igh-Frequency \textbf{R}esidual \textbf{R}efinement (HRR) module, which adopts the diffusion models to predict the residual (high-frequency information, including text edges), between the ground-truth and the CP-predicted image. DocDiff is a compact and computationally efficient model that benefits from a well-designed network architecture, an optimized training loss objective, and a deterministic sampling process with short time steps. Extensive experiments demonstrate that DocDiff achieves state-of-the-art (SOTA) performance on multiple benchmark datasets, and can significantly enhance the readability and recognizability of degraded document images. Furthermore, our proposed HRR module in pre-trained DocDiff is plug-and-play and ready-to-use, with only 4.17M parameters. It greatly sharpens the text edges generated by SOTA deblurring methods without additional joint training. \textbf{Available codes}: \href{https://github.com/Royalvice/DocDiff}{https://github.com/Royalvice/DocDiff}.
\end{abstract}

%%
%% The code below is generated by the tool at http://dl.acm.org/ccs.cfm.
%% Please copy and paste the code instead of the example below.
%%


%%
%% Keywords. The author(s) should pick words that accurately describe
%% the work being presented. Separate the keywords with commas.

%% A "teaser" image appears between the author and affiliation
%% information and the body of the document, and typically spans the
%% page.
% \begin{teaserfigure}
%   \includegraphics[width=\textwidth]{sampleteaser}
%   \caption{Seattle Mariners at Spring Training, 2010.}
%   \Description{Enjoying the baseball game from the third-base
%   seats. Ichiro Suzuki preparing to bat.}
%   \label{fig:teaser}
% \end{teaserfigure}

% \received{20 February 2007}
% \received[revised]{12 March 2009}
% \received[accepted]{5 June 2009}

%%
%% This command processes the author and affiliation and title
%% information and builds the first part of the formatted document.
\maketitle

\section{Introduction}

Document images are widely used in multi-media applications. The vulnerability to degradation is one of the challenges encountered in processing such highly structured data that differs significantly in pixel distribution from natural scene images. As an important pre-processing step, document enhancement aims to restore degraded document images to improve their readability and the performance of OCR systems \cite{KhamekhemJemni2022}. In this paper, we focus on three major document enhancement tasks: document deblurring, document denoising and binarization, i.e., to remove fragmented noise from documents, such as smears and bleed-throughs, and watermark and seal removal. (Figure \ref{fig:noise} in Appendix shows several degraded document images addressed in this paper. )

\begin{figure}[!t]
    \centering
    \captionsetup[subfloat]{labelsep=none, format=plain, labelformat=empty}
    \subfloat[\textbf{Blurred text lines}]{\includegraphics[width=0.52\linewidth]{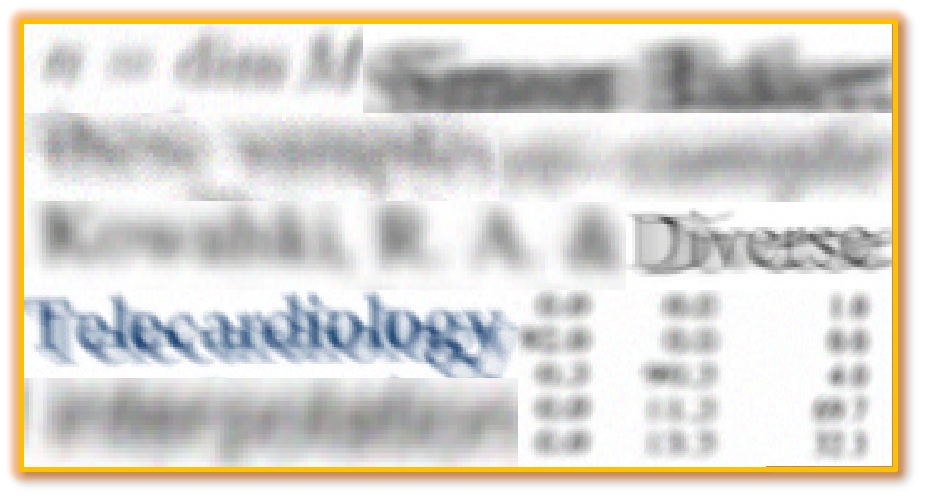}}
    \subfloat[\textbf{Ground-truth}]{\includegraphics[width=0.52\linewidth]{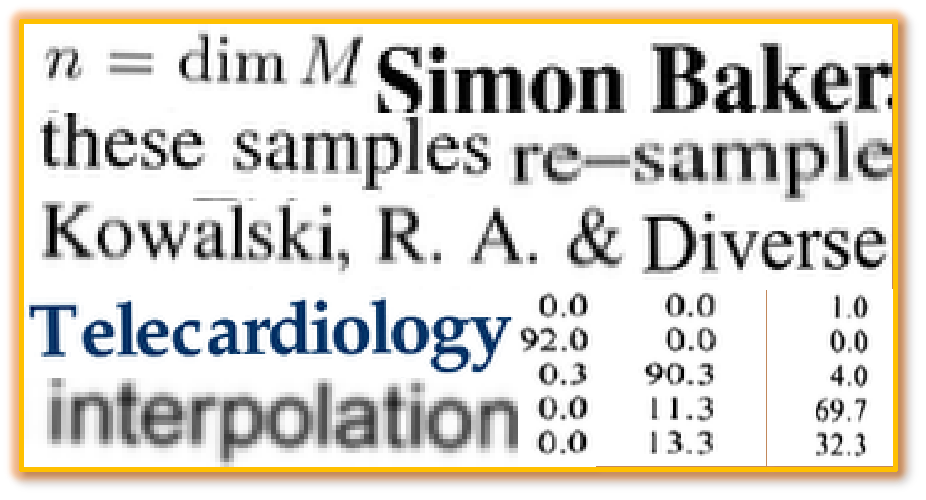}}
    \quad
    \subfloat[\textbf{DocDiff}]{\includegraphics[width=0.21\linewidth, height = 0.55\linewidth]{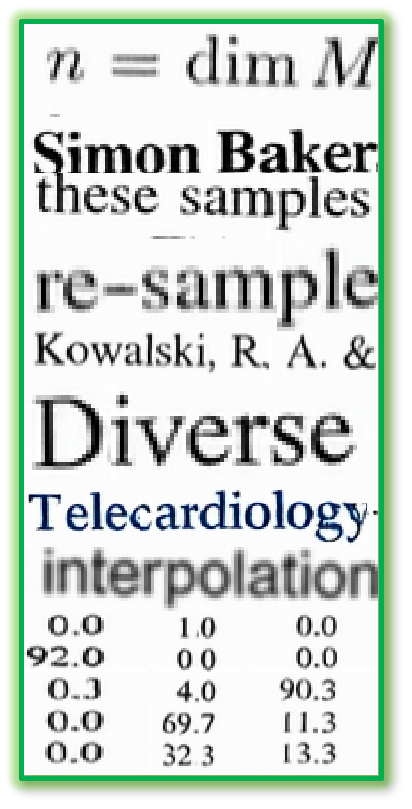}}
    \subfloat[HINet \cite{chen2021hinet} / HINet+\textbf{HRR}]{\includegraphics[width=0.42\linewidth, height = 0.55\linewidth]{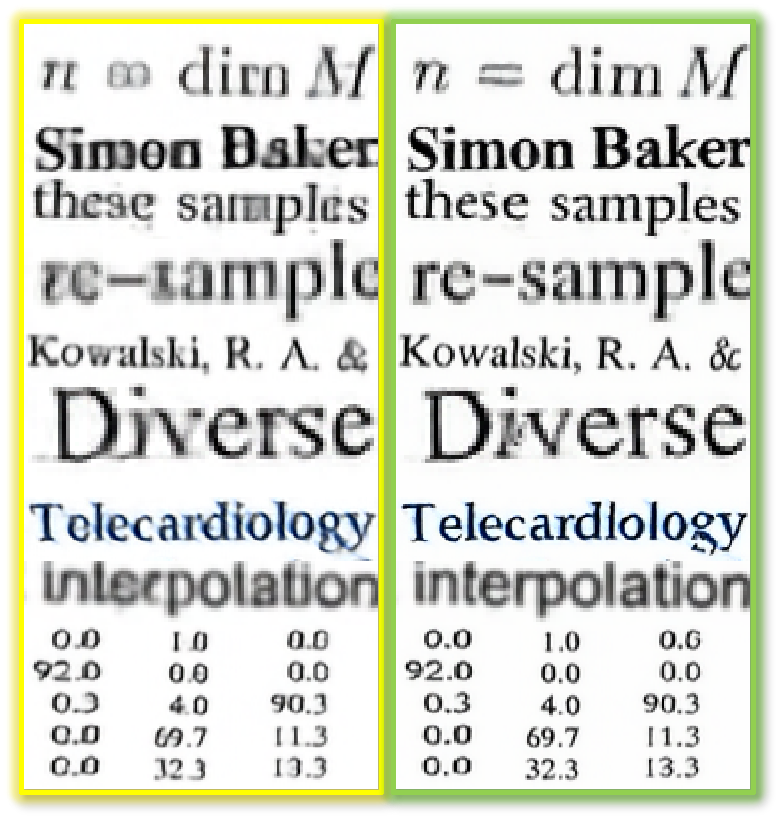}}
    \subfloat[DE-GAN \cite{souibgui2020gan} / DEGAN+\textbf{HRR}]{\includegraphics[width=0.42\linewidth, height = 0.55\linewidth]{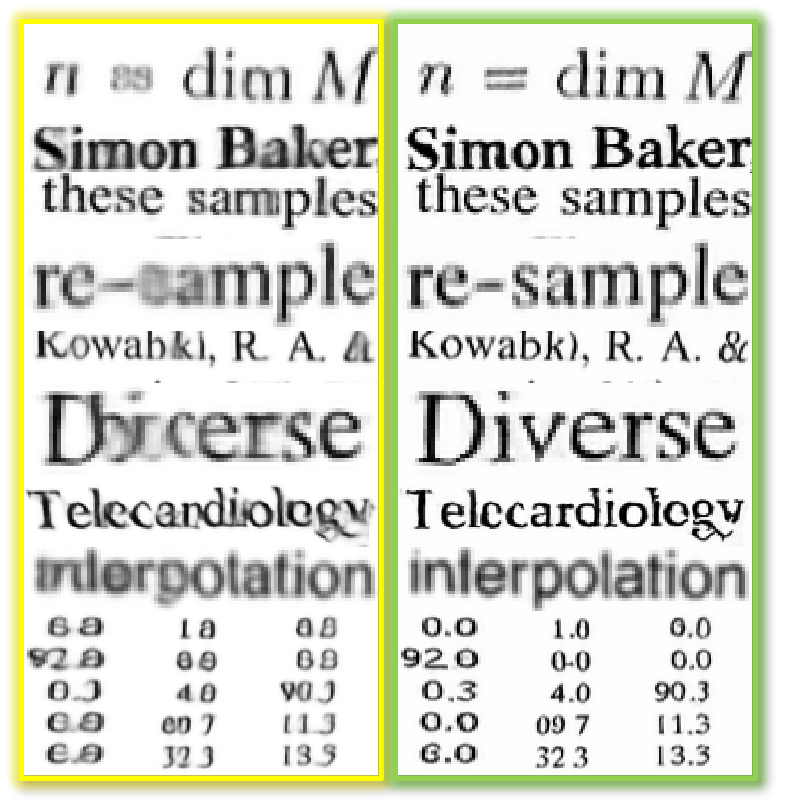}}
    \caption{Compared to DE-GAN \cite{souibgui2020gan} and HINet \cite{chen2021hinet}, DocDiff generates sharper text edges. Our proposed HRR module effectively mitigates the problem of distorted and blurred characters generated by regression-based methods, regardless of whether they are designed for natural or document scenes. Note that the HRR module does not require additional joint training with DE-GAN \cite{souibgui2020gan} and HINet \cite{chen2021hinet}. Its weights are derived from the pre-trained DocDiff.}
    \label{fig:kaipian}
\end{figure}

The major challenges for document enhancement are noise elimination and pixel-level text generation with low latency on high resolution document images. Specially, the presence of diverse noise types in document images, comprising of both global noise such as blurring and local noise such as smears, bleed-throughs, and seals, along with their potential combination, poses a significant challenge for noise elimination. Moreover, the task of generating text-laden images is fraught with difficulty. Unlike images that depict natural scenes, the high-frequency information of text-laden images is mostly concentrated on the text edges. The slightest erroneous pixel modifications at the text edges have the potential to alter the semantic meaning of a character, rendering it illegible or unrecognizable by OCR systems. %We emphasize the importance of character consistency between the enhanced and original images for document enhancement. 
Thus, document enhancement does not prioritize generation diversity, which differs from the pursuit of recovering multiple distinct denoised images in natural scenes \cite{whang2022deblurring, shang2023resdiff}. In practice, a typical document image contains millons of pixels. To ensure the efficiency of the entire document analysis system, pre-processing speed is crucial, which demands models to be as lightweight as possible.
%Removing noise or improving image quality at the expense of altering characters is counterproductive, as it defeats the primary objectives of enhancing the readability and OCR performance.

% \begin{figure}[!t]
%     \centering
%     \subfloat[Various noises (Broken holes, Bleed-through, Smear, Faint ink)]{\includegraphics[width=\linewidth]{img/noise.png}}
%     \quad
%     \subfloat[Blur]{\includegraphics[width=0.5\linewidth]{img/blur.png}}
%     \subfloat[Watermark]{\includegraphics[width=0.25\linewidth]{img/watermark.png}}
%     \subfloat[Seal]{\includegraphics[width=0.25\linewidth]{img/seal.jpg}}
%     \caption{Examples of degraded document images.}
%     \label{fig:noise}
% \end{figure}
Currently, existing document enhancement methods \cite{souibgui2020gan, souibgui2022docentr, Suh2022} are deep learning-based regression methods. Due to the problem of "regression to the mean", these methods \cite{souibgui2020gan, souibgui2022docentr, Suh2022} optimized with pixel-level loss produce blurry and distorted text edges. Additionally, due to the existence of numerous non-text regions in high-resolution document images, GAN-based methods \cite{souibgui2020gan} are prone to mode collapse when trained on local patches. In natural scenes, many diffusion-based methods \cite{ saharia2022image, whang2022deblurring} try to restore degraded images with more details. However, there are challenges in directly applying these methods to document enhancement. Foremost, their high training costs and excessively long sampling steps make them difficult to be practically implemented in document analysis systems. For these methods, shorter inference time implies the need for shorter sampling steps, which can lead to substantial performance degradation. Besides, the generation diversity of these methods can result in character inconsistency between the conditions and the sampled images. 
% During inference, these methods \cite{souibgui2020gan, Zhao2019, Suh2022, souibgui2022docentr, yang2023novel} require the crop-predict-merge strategy to achieve better performance, which decreases the processing speed of high-resolution document images under limited memory.

Considering these problems, we transform document enhancement into the task of conditional image-to-image generation and propose DocDiff, which consists of two modules: the Coarse Predictor (CP) and the High-Frequency Residual Refinement (HRR) module. The CP takes degraded document images as input and approximately restores their clean versions. The HRR module leverages diffusion models to sharp the text edges produced by CP accurately and efficiently. To avoid enhancing the generation quality at the expense of altering the characters and speed the inference, we adjust the optimization objective during training and adopt a short-step deterministic sampling strategy during inference. Specifically, we allow the HRR module to learn the distribution of residuals between the ground-truth images and the CP-predicted images (conditions). While ensuring consistency in the reverse diffusion process, we enable the HRR to directly predict the original residual rather than the added noise. This allows DocDiff to produce reasonable images that are highly correlated with the conditions in the first few steps of the reverse diffusion, based on the premise of using a channel-wise concatenation conditioning scheme \cite{whang2022deblurring, saharia2022palette, saharia2022image}. While sacrificing generation diversity, which is not a critical factor for document enhancement, this operation considerably reduces the number of diffusion steps required for sampling and results in a reduced range of potential outputs generated by the conditional diffusion model, which effectively mitigates the production of distorted text edges and the replacement of characters. Overall, DocDiff undergoes end-to-end training with frequency separation through the joint use of pixel loss and modified diffusion model loss and applies the deterministic short-step sampling. DocDiff strikes the balance between innovation and practicality, achieving outstanding performance with a tiny and efficient network that contains only 8M parameters.

Experimental results on the three benchmark datasets (Document Deblurring \cite{hradivs2015convolutional} and (H-)DIBCO \cite{Pratikakis2018, Pratikakis2019}) demonstrate that DocDiff achieves SOTA performance in terms of perceptual quality for low-level deblurring task and competitive performance for high-level binarization task. More importantly, DocDiff achieves competitive deblurring performance with only 5 sampling steps. For the task of watermark and seal removal, we generated paired datasets using in-house document images. Experimental results demonstrate that DocDiff can effectively remove watermarks and seals while preserving the covered characters. Specifically, for seal removal, DocDiff trained on the synthesized dataset shows promising performance in real-world scenarios. Ablation experiments demonstrate the effectiveness of the HRR module on sharpening blurred characters generated by regression methods, as shown in Fig. ~\ref{fig:kaipian}.

In summary, the contributions of our paper are as follows:
\begin{itemize}
    \setlength{\itemsep}{0pt}
    \setlength{\parsep}{0pt}
    \setlength{\parskip}{0pt}
  \item We present a novel framework, named DocDiff. To the best of our knowledge, it is the first diffusion-based method which is specifically designed for diverse chanllenging document enhancement tasks.
  \item We propose a plug-and-play High-Frequency Residual Refinement (HRR) module to refine the generation of text edges. We demonstrate that the HRR module is capable of directly and substantially enhancing the perceptual quality of deblurred images generated by regression methods without requiring any additional training. 
  \item DocDiff is a tiny, flexible, efficient and train-stable generative model. Our experiments show that DocDiff achieves competitive performance with only 5 sampling steps. Compared with non-diffusion-based methods \cite{souibgui2020gan, Zhao2019, KhamekhemJemni2022, souibgui2022docentr}, DocDiff's inference is fast with the same level of performance. Additionally, DocDiff is trained inexpensively, avoids mode collapse and can enhance both handwritten and printed document images at any resolution. 
  %This is because DocDiff can enhance the document image at the original resolution, while the patch-based approach has to crop and then enhance and then stitch from the original image.
  \item Adequate ablation studies and comparative experiments show that DocDiff achieves competitive performance on the tasks of deblurring, denoising, watermark and seal removal on documents. Our results highlight the benefits of various components of DocDiff, which collectively contribute to its superior performance.
\end{itemize}

\section{Related works}
\subsection{Document Enhancement}

The pixel distribution in document images differs significantly from that of natural scene images, and most document images have a resolution greater than 1k. Therefore, it is crucial to develop specialized enhancement models for document scenarios to handle the degradation of different types of documents efficiently and robustly. The currently popular document enhancement methods \cite{lin2020bedsr,souibgui2020gan,souibgui2022docentr,KhamekhemJemni2022} are predominantly based on deep learning regression metohds. These methods aim to achieve higher PSNR by minimizing $\mathcal{L}_1$ or $\mathcal{L}_2$ pixel loss. However, distortion metrics like PSNR only partially align with human perception \cite{blau2018perception}. This problem is particularly noticeable in document scenarios (see details depicted in Fig.~\ref{fig:metrics} in Appendix). Although GAN-based methods \cite{souibgui2020gan, Suh2022, Zhao2019} utilize a combination of content and adversarial losses to generate images with sharp edges, training GANs on high-resolution document datasets can be challenging because the cropped patches typically have a significant number of identical patterns, which increases the risk of mode collapse. 

\begin{figure*}[!htbp]
\centering
\includegraphics[width=0.65\linewidth, height=0.55\linewidth]{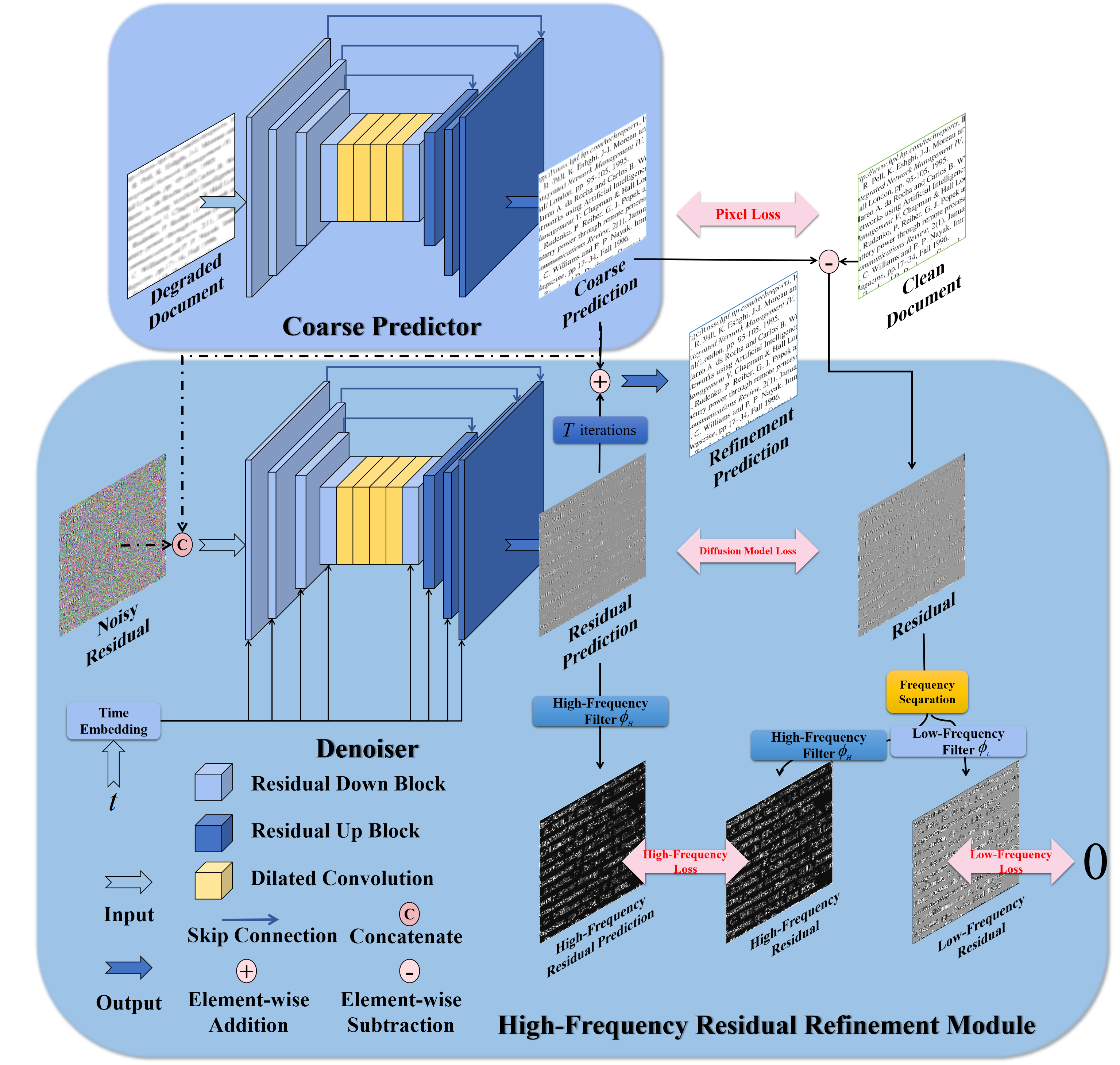}%
\centering
\caption{A overview of the proposed DocDiff for document enhancement. Take document deblurring as an example.}
\label{img:model}
\centering
\end{figure*}

\subsection{Diffusion-based Image-to-image}

Recently, Diffusion Probabilistic Models (DPMs) \cite{ho2020denoising, song2020denoising} have been widely adopted for conditional image generation \cite{saharia2022image, saharia2022palette, rombach2022high, whang2022deblurring, shang2023resdiff, li2022srdiff, niu2023cdpmsr, Wang_2023_CVPR, wang2022zero}.
Saharia et al. \cite{saharia2022image} present SR3, which adapts diffusion models to image super-resolution.
Saharia et al. \cite{saharia2022palette} propose Palette, which is a multi-task image-to-image diffusion model. Palette has demonstrated the excellent performance of diffusion models in the field of conditional image generation, including colorization, inpainting and JPEG restoration.
% Rombach et al. \cite{rombach2022high} train the diffusion models on the latent space that significantly lower computational costs and introduce the cross-attention conditioning mechanism that allow multimodal learning.
In \cite{whang2022deblurring, shang2023resdiff, li2022srdiff}, they all utilize the prediction-refinement framework where the diffusion models are used to predict the residual. Different from \cite{shang2023resdiff, li2022srdiff, niu2023cdpmsr}, DocDiff is trained end-to-end. 

Although effective for denoising natural images, they are not specifically designed for document images. Methods \cite{saharia2022image, saharia2022palette, rombach2022high, shang2023resdiff, li2022srdiff, niu2023cdpmsr} cannot handle images of arbitrary resolution. Due to their large networks, long sampling steps and large number of cropped patches, these lead to prohibitively long inference time on high resolution document images. Although Method \cite{whang2022deblurring} is capable of processing images of any resolution and has a relatively small network structure, its training method of predicting noise and stochastic sampling strategy produce diverse characters and distorted text edges, which is not suitable for document enhancement.

\section{METHODOLOGY}
The overall architecture of DocDiff is shown in Fig. \ref{img:model}. DocDiff consists of two modules: the Coarse Predictor (CP) and the High-Frequency Residual Refinement (HRR) module. Due to the fixed pattern of document images and their varying resolutions, we adopt a compact U-Net structure for the CP and HRR module, modified from \cite{ho2020denoising}. We replace the self-attention layer in the middle with four layers of dilated convolutions to increase the receptive field, and remove the remaining self-attention and normalization layers. To reduce computational complexity, we compress the parameters of the CP and HRR module to 4.03M and 4.17M respectively, while ensuring performance, making them much smaller than existing document enhancement methods \cite{Zhao2019, souibgui2020gan, souibgui2022docentr, yang2023novel}. Overrall, DocDiff is a fully convolutional model that employs a combination of pixel loss and diffusion model loss, facilitating end-to-end training with frequency separation.

\subsection{Coarse Predictor}

The objective of the Coarse Predictor $C_{\theta}$ is to approximately restore a degraded document image $y$ into its clean version $x_{\rm gt}$ at the pixel level. The $\mathcal{L}_{\rm pixel}$ can be defined as the mean square error between the coarse prediction $x^C$ and the $x_{\rm gt}$:
\begin{equation}
     x^C = C_{\theta}(y)
\end{equation}
\begin{equation}
    \label{eq:cp}
    \mathcal{L}_{\rm pixel} = \mathbb{E} \| x^C - x_{\rm gt} \|_{2}
\end{equation}

During text pixel generation, the Coarse Predictor can effectively restore the primary content of the text, but it may not be able to accurately capture the high-frequency information in the text edges. This leads to significant blurring at the edges of the text. As shown in Fig.~\ref{fig:kaipian}, this is a well-known limitation of CNN-based regression methods due to the problem of "regression to the mean". It is challenging to address this issue by simply cascading more convolutional layers.

\subsection{High-Frequency Residual Refinement Module}

To address the problem mentioned above, we introduce the High-Frequency Residual Refinement (HRR) module, which is capable of generating samples from the learned posterior distribution. The core component of HRR module id a Denoiser $f_\theta$ which leverages DPMs to estimate the distribution of residuals between the ground-truth images and the images generated by the CP. In contrast to prior research \cite{li2022srdiff, shang2023resdiff, niu2023cdpmsr}, we design the HRR module to address not only the "regression to the mean" flaw found in a single regression model (in this case, the CP), but also to be effective across a variety of regression methods. To this end, We perform end-to-end joint training of both CP and HRR modules, rather than separately. By this way, the HRR module can dynamically adjust and capture more patterns. Extensive experiments show that this training strategy effectively enhances the sharpness of characters generated by different regression-based deblurring methods \cite{kupyn2019deblurgan, souibgui2020gan, souibgui2022docentr, chen2021hinet, zamir2021multi}, without requiring joint training like \cite{niu2023cdpmsr, shang2023resdiff}. 

\subsubsection{Denoiser $f_\theta$}
\label{sec:rdm}
Followed by \cite{ho2020denoising, song2020denoising}, the HRR module executes the \textbf{forward noise-adding process} and the \textbf{reverse denoising process} to model the residual distributions.

\textbf{Forward noise-adding process:} Given the clean document image $x_{\rm gt}$ and its approximate estimate $x^C$, we calculate their residuals $x_{\rm res}$. We assign $x_0$ as $x_{\rm res}$, and then sequentially introduce Gaussian noise based on the time step, as follows:

\begin{equation}
    x_0 = x_{\rm res} = x_{\rm gt}-x^C
\end{equation}

\begin{equation}
q\left(x_t \mid x_{t-1}\right) = \mathcal{N}\left(x_t ; \sqrt{\alpha_t} x_{t-1},\left(1-\alpha_t\right) \mathbf{I}\right)
\end{equation}

\begin{equation}
q\left(x_{1: T} \mid x_0\right)=\prod_{t=1}^T q\left(x_t \mid x_{t-1}\right)
\end{equation}

\begin{equation}
q\left(x_t \mid x_0\right) = \mathcal{N}\left(x_t ; \sqrt{\bar{\alpha}_t} x_0,\left(1-\bar{\alpha}_t\right) \mathbf{I}\right)
\end{equation}
where $\alpha_t$ is a hyperparameter between 0 and 1 that controls the variance of the added Gaussian noise at each time step, for all $t=1,...,T$, and $\alpha_0 = 1, \bar{\alpha}_t = \prod_{i=0}^t \alpha_i$. There are no learnable parameters in the forward process, and $x_{1: T}$ have the same size as $x_0$. With the reparameterization trick, $x_t$ can be written as:

\begin{equation}
\label{eq:tran}
x_t = \sqrt{\bar{\alpha}_t} x_0+\sqrt{1-\bar{\alpha}_t} \epsilon, \epsilon \sim \mathcal{N}(\mathbf{0}, \mathbf{I})
\end{equation}

\textbf{Reverse denoising process:} The reverse process transforms Gaussian noise back into the residual distributions conditioned on $x^C$. We can write the reverse diffusion step:

\begin{equation}
    \label{eq:8}
q\left(x_{t-1} \mid x_t, x_0\right)=\mathcal{N}\left(x_{t-1} ; \mu_t(x_t, x_0), \beta_t(x_t, x_0) \mathbf{I}\right)
\end{equation}
where $\mu_t(x_t, x_0)$ and $\beta_t(x_t, x_0)$ are the mean and variance, respectively. Followed by \cite{song2020denoising}, we perform the deterministic reverse process $q\left(x_{t-1} \mid x_t, x_0\right)$ with zero variance and the mean can be computed:
\begin{equation}
    \label{eq:9}
    \mu_t(x_t, x_0) = \sqrt{\bar{\alpha}_{t-1}} x_0+\sqrt{1-\bar{\alpha}_{t-1}} \cdot \underbrace{\frac{x_t-\sqrt{\bar{\alpha}_t} x_0}{\sqrt{1-\bar{\alpha}_t}}}_\epsilon
\end{equation}
\begin{equation}
    \label{eq:10}
    \beta_t(x_t, x_0) = 0
\end{equation}

Given the Denoiser $f_\theta$, the posterior $q\left(x_{t-1} \mid x_t, x_0\right)$ can be parameterized:
\begin{equation}
    \label{eq:11}
    p_\theta(x_{t-1} \mid x_t, x^C) = q(x_{t-1} \mid x_t, f_\theta(x_t, t, x^C))
\end{equation}

We emphasize the significance of the condition $x^C$ in the conditional distribution $p_\theta(x_{t-1} \mid x_t, x^C)$. At each time step, it is crucial to sample the residual that closely relate to $x^C$ at the pixel level of the characters. 

The Denoiser $f_\theta$ can be trained to predict the original data $x_0$ or the added noise $\epsilon$. To increase the diversity of generated natural images, existing methods \cite{ho2020denoising, whang2022deblurring, saharia2022palette, saharia2022image} typically predict the added noise $\epsilon$. The prediction of $x_0$ and $\epsilon$ is equivalent in unconditional generation, as they can be transformed into each other through Eq.\ref{eq:tran}. However, for conditional generation, predicting $x_0$ and predicting $\epsilon$ are not equivalent under the premise of using a channel-wise concatenation conditioning scheme \cite{whang2022deblurring, saharia2022palette, saharia2022image} to introduce the condition $x^C$. When predicting $\epsilon$, the denoiser can only learn from the noisy $x_t$. But when predicting $x_0$, the denoiser can also learn from the conditional channels ($x^C$). This sacrifices diversity but significantly improves the generation quality of the first few steps in the reverse process. To this end, we train the Denoiser $f_\theta$ to directly predict $x_0$, which aligns with the goals of document enhancement. The training objective is to minimize the distance between $p_\theta(x_{t-1} \mid x_t, x^C)$ and the true posterior $q\left(x_{t-1} \mid x_t, x_0\right)$:

\begin{equation}
        \mathcal{L}_{DM} = \mathbb{E}\left\|x_{0}-f_\theta\left(\sqrt{\bar{\alpha}_t}x_{\rm res}+\sqrt{1-\bar{\alpha}_t} \epsilon, t, \hat{x}^C\right)\right\|_2
\end{equation}
where $\hat{x}^C$ is a clone of $x^C$ in memory and does not participate in gradient calculations. Exactly, the gradient from the loss only flows through $x_{\rm res}$ from $f_\theta$ to $C_\theta$.

Given the trained $C_\theta$ and $f_\theta$, integrating the above equation we finally have the deterministic reverse process:

\begin{equation}
    \hat{x}_{\rm res} = f_\theta(x_t, t, C_\theta(y))
\end{equation}

\begin{equation}
    x_{t-1} = \sqrt{\bar{\alpha}_{t-1}} \hat{x}_{\rm res}+\sqrt{1-\bar{\alpha}_{t-1}} \cdot \frac{x_t-\sqrt{\bar{\alpha}_t} \hat{x}_{\rm res}}{\sqrt{1-\bar{\alpha}_t}}
\end{equation}

\subsubsection{Frequency Separation Training}

Numerous prior studies \cite{fritsche2019frequency, fuoli2021fourier, shang2023resdiff} have demonstrated that processing high and low frequency information separately enhances the quality and level of detail in generated images of natural scenes. To further refine the generation quality, DocDiff is trained with frequency separation. Specifically, we employ simple linear filters to separate the residuals into the low and high frequencies. As the spatial-domain addition is equivalent to the frequency-domain addition, frequency separation can be written directly as:
\begin{equation}
\label{eq:hl}
    x = \phi_{\rm L} \ast x + \phi_{\rm H} \ast x
\end{equation}

\begin{table*}[!htbp]
\caption{Quantitative ablation study results on the Document Deblurring Dataset \cite{hradivs2015convolutional}. Best values are highlighted in \textcolor{red}{red}, second best are highlighted in \textcolor{blue}{blue}. (CP: Coarse Predictor, CR: Cascade Refinement Module, HRR: High-Frequency Residual Refinement Module, EMA: Exponential Moving Average, FS: Frequency Separation Training.)}
\label{tab:abl}
\resizebox{\linewidth}{!}{%
\begin{tabular}{llllllllllllllll}
\hline
\multicolumn{10}{c}{Method} & \multicolumn{4}{c}{\textbf{Perceptual}} & \multicolumn{2}{c}{\textbf{Distortion}} \\ \hline
\multicolumn{3}{c}{\begin{tabular}[c]{@{}c@{}}Structure\\ components\end{tabular}} & \multicolumn{2}{c}{\begin{tabular}[c]{@{}c@{}}Training\\ techniques\end{tabular}} & \multicolumn{2}{c}{Resolutions} & \multicolumn{2}{c}{Sampling} & Parameters & \multicolumn{2}{c}{NR} & \multicolumn{2}{c}{FR} & \multicolumn{2}{c}{} \\ \hline
CP & CR & HRR & EMA & FS & Non-native & Native & Stochastic & Deterministic &  & MANIQA↑ & MUSIQ↑ & DISTS↓ & LPIPS↓ & PSNR↑ & SSIM↑ \\ \hline
{$\checkmark$} &  &  & {$\checkmark$} &  &  & {$\checkmark$} & - & - & 4.03M & 0.6525 & 46.15 & 0.0951 & 0.0766 & \textcolor{blue}{24.66} & \textcolor{blue}{0.9574} \\
{$\checkmark$} & {$\checkmark$} &  & {$\checkmark$} & {$\checkmark$} &  & {$\checkmark$} & - & - & 8.06M & 0.6584 & 45.39 & 0.0688 & 0.0824 & \textcolor{red}{24.74} & \textcolor{red}{0.9610} \\
{$\checkmark$} &  & {$\checkmark$} &  &  & {$\checkmark$} &  &  & {$\checkmark$} & 8.20M & 0.6900 & 50.16 & 0.0671 & 0.0492 & 20.98 & 0.9025 \\
{$\checkmark$} &  & {$\checkmark$} & {$\checkmark$} &  & {$\checkmark$} &  &  & {$\checkmark$} & 8.20M & 0.6917 & 50.21 & 0.0648 & 0.0499 & 20.43 & 0.8998 \\
{$\checkmark$} &  & {$\checkmark$} & {$\checkmark$} & {$\checkmark$} & {$\checkmark$} &  & {$\checkmark$} &  & 8.20M & 0.6706 & 50.05 & 0.1778 & 0.1481 & 18.72 & 0.8507 \\
{$\checkmark$} &  & {$\checkmark$} & {$\checkmark$} & {$\checkmark$} & {$\checkmark$} &  &  & {$\checkmark$} & 8.20M & \textcolor{blue}{0.6971} & \textcolor{blue}{50.31} & \textcolor{blue}{0.0636} & \textcolor{blue}{0.0474} & 20.46 & 0.9006 \\
{$\checkmark$} &  & {$\checkmark$} & {$\checkmark$} & {$\checkmark$} &  & {$\checkmark$} &  & {$\checkmark$} & 8.20M & \textcolor{red}{0.7174} & \textcolor{red}{50.62} & \textcolor{red}{0.0611} & \textcolor{red}{0.0307} & 23.28 & 0.9505 \\ \hline
\end{tabular}%
}
\end{table*}

where $\phi_{\rm L}$ is the low-pass filter and $\phi_{\rm H}$ is the high-pass filter. In pratice, to extract residuals along the text edges, we commonly set $\phi_{\rm H}$ as the Laplacian kernel. The low-frequency information is obtained according to Eq.\ref{eq:hl}. Our approach differs from \cite{wu2022medsegdiff, shang2023resdiff} in that it does not require performing the Fast Fourier Transform (FFT) and parameterizing frequency separation in the frequency domain. The high-frequency information in document images is primarily concentrated at the text edges. Leveraging this prior knowledge by using the Laplacian kernel as a high-pass filter not only simplifies training time consumption but also proves to be highly effective.

Our \textbf{goal} is to maximize the capacity of the Denoiser $f_\theta$ to restore the missing high-frequency information in the Coarse Predictor $C_\theta$ prediction, while minimizing the task burden of $f_\theta$ through the support of $C_\theta$. In this perspective, both $C_\theta$ and $f_\theta$ necessitate the restoration of distinct high and low frequencies, yet they exhibit dissimilar specializations. Specifically, $C_\theta$ predominantly reconstructs low-frequency information, while $f_\theta$ specializes in restoring high-frequency details:
\begin{equation}
        \mathcal{L}^{\rm low} = \mathbb{E} \| \phi_{\rm L} \ast (x^C - x_{\rm gt}) \|_{2}
\end{equation}
\begin{equation}
    \mathcal{L}^{\rm high} = \mathbb{E}\left\|\phi_{\rm H} \ast \left(x_{0} - f_\theta\left(x_t, t, \hat{x}^C\right)\right)\right\|_2
\end{equation}
Thus, the combined loss for $C_\theta$ and $f_\theta$, together with the overall loss for DocDiff, are given by:
\begin{equation}
    \label{low}
        \mathcal{L}^{\rm low}_{\rm pixel} = \mathcal{L}_{\rm pixel} + \beta_0 \mathcal{L}^{\rm low}
\end{equation}
\begin{equation}
    \label{high}
        \mathcal{L}^{\rm high}_{DM} = \mathcal{L}_{\rm DM} + \beta_0 \mathcal{L}^{\rm high}
\end{equation}
\begin{equation}
    \label{loss:total}
    \mathcal{L}_{total} = \beta_1\mathcal{L}^{\rm low}_{\rm pixel} + \mathcal{L}^{\rm high}_{DM}
\end{equation}
Algorithm \ref{alg:docdiff} outlines the complete training and inference processes of DocDiff.
\begin{algorithm}[!h]
\caption{DocDiff}
\label{alg:docdiff}
\small
  \SetKwInOut{Training}{Training}\SetKwInOut{Sampling}{Sampling}
  \SetKwData{Input}{Input:}\SetKwFunction{Uniform}{Uniform}
  \SetKwInOut{Require}{Require}
  \raggedright
  \Require {$C_{\theta}$: Coarse Predictor, $f_{\theta}$: Denoiser, \\$y$: Degraded document image, $\phi_{\rm H}$: High frequency filter, \\ $x_{\rm gt}$: Clean document image, $\alpha_{0:T}$: Noise schedule.}
  \Training{}
\begin{algorithmic}[1]
    \REPEAT
      \STATE $(y, x_{\rm gt}) \sim q(y, x_{\rm gt})$, $t \sim$ Uniform{$\{1,...,T\}$}, $\epsilon \sim \mathcal{N}(\mathbf{0},\mathbf{I} )$
      \STATE $x^{C} = C_{\theta}(y)$ \hfill $\triangleright$ Coarse prediction
      \STATE $x_{t} = \sqrt{\bar{\alpha}}_{t}(x_{\rm gt} - x^{C}) + \sqrt{1-\bar{\alpha}}_{t}\epsilon$ \hfill $\triangleright$ Forward diffusion
      \STATE $\hat{x}_{\rm res} = f_{\theta}(x_{t}, t, x^{C})$ \hfill $\triangleright$ Residual prediction
      \STATE Take a gradient descent step on\\
       $\mathcal{L}_{total}(x^{C}, \hat{x}_{\rm res}, x_{\rm gt}, \phi_{\rm H} ; C_{\theta}, f_{\theta})$ \\
       \hfill $\triangleright$ Frequency separation training; see Eq. \ref{loss:total}
    \UNTIL{converged}
\end{algorithmic}

\Sampling{}
\begin{algorithmic}[1]
  \STATE $x^{C} = C_{\theta}(y)$ \hfill $\triangleright$ Coarse prediction
  \STATE $x_{T} \sim \mathcal{N}(\mathbf{0},\mathbf{I} )$
  \FOR{$t=T,...,1$}
    \STATE $\hat{x}_{\rm res} = f_\theta(x_t, t, x^{C})$
    \STATE $x_{t-1} = \sqrt{\bar{\alpha}_{t-1}} \hat{x}_{\rm res} + \frac{\sqrt{1-\bar{\alpha}_{t-1}}(x_{t} - \sqrt{\bar{\alpha}_{t}} \hat{x}_{\rm res})}{\sqrt{1-\bar{\alpha}_{t}}}$ \\
    \hfill $\triangleright$ Deterministic reverse diffusion
  \ENDFOR
  \RETURN{$x^C + x_{0}$} \hfill $\triangleright$ Return high-frequency residual refinement
\end{algorithmic}
\end{algorithm}
\section{Experiments and Results}
\subsection{Datasets and Implementation details}
To address various document enhancement tasks, we train and evaluate our models on distinct datasets. For a fair comparison, we use the source codes provided by the authors and follow the same experimental settings.

\textbf{Document Deblurring:} We train and evaluate DocDiff on the widely-used Document Deblurring Dataset \cite{hradivs2015convolutional}, which includes 66,000 pairs of clean and blurry 300 × 300 patches extracted from diverse pages of different documents. Each blur kernel is distinct. We randomly select 30,000 patches for training and 10,000 patches for testing. 

\textbf{Document Denoising and Binarization:} We evaluate DocDiff on two of the most challenging datasets from the annual (Handwritten) Document Image Binarization Competition ((H-)DIBCO) \cite{Gatos2009, Pratikakis2010, Pratikakis2011, Pratikakis2012, Pratikakis2013, Ntirogiannis2014a, Pratikakis2016, Pratikakis2017, Pratikakis2018, Pratikakis2019}: H-DIBCO‘18 \cite{Pratikakis2018} and DIBCO‘19 \cite{Pratikakis2019}. Followed by \cite{yang2023novel, Suh2022, yang2023gdb}, the training set includes the remaining years of DIBCO datasets \cite{Gatos2009, Pratikakis2010, Pratikakis2011, Pratikakis2012, Pratikakis2013, Ntirogiannis2014a, Pratikakis2016, Pratikakis2017}, the Noisy Office Database \cite{Zamora-Martinez2007}, the Bickley Diary dataset \cite{Deng2010}, Persian Heritage Image Binarization Dataset (PHIDB) \cite{Nafchi2013}, and the Synchromedia Multispectral dataset (S-MS) \cite{Hedjam2015}. 

\begin{figure*}[!t]
    \centering
    \captionsetup[subfloat]{labelsep=none, format=plain, labelformat=empty}
    \subfloat[\textbf{Input}]{\includegraphics[width=0.14\linewidth, height=0.2\linewidth]{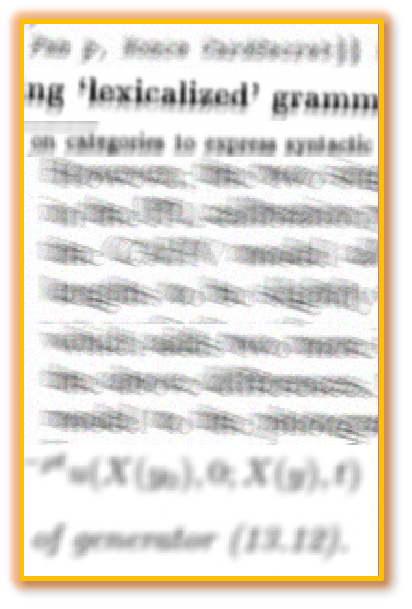}}
    \subfloat[\textbf{Ground-truth}]{\includegraphics[width=0.14\linewidth, height=0.2\linewidth]{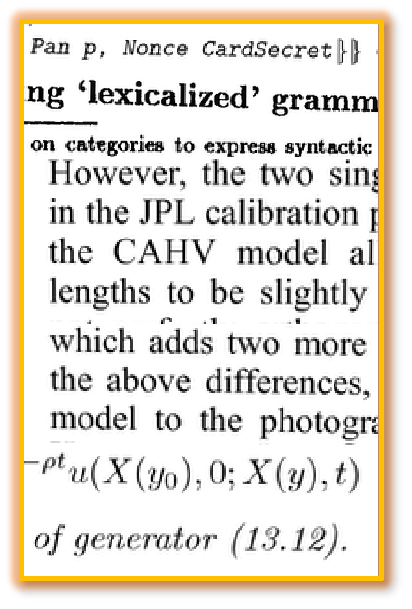}}
    \subfloat[DE-GAN \cite{souibgui2020gan} \quad \quad \quad DEGAN+\textbf{HRR}]{\includegraphics[width=0.29\linewidth, height=0.2\linewidth]{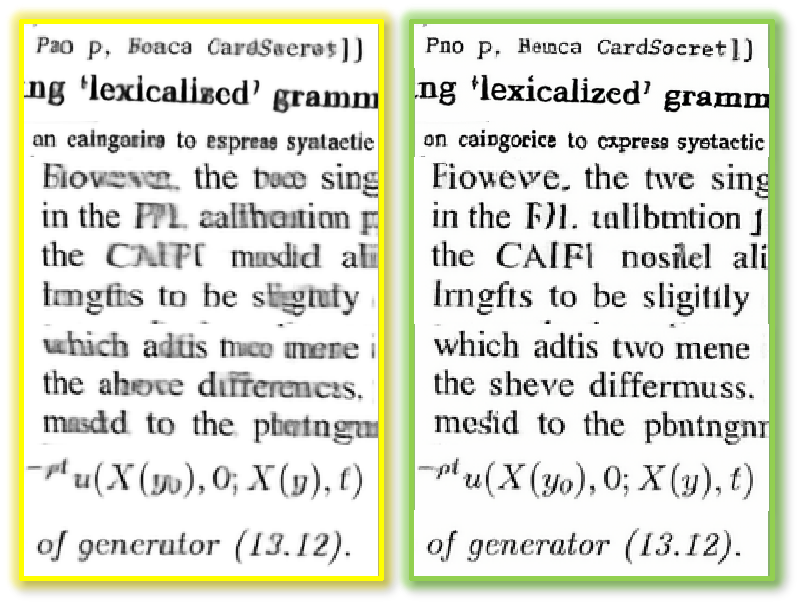}}
    \subfloat[HINet \cite{chen2021hinet} \quad \quad \quad \quad \quad HINet+\textbf{HRR}]{\includegraphics[width=0.29\linewidth, height=0.2\linewidth]{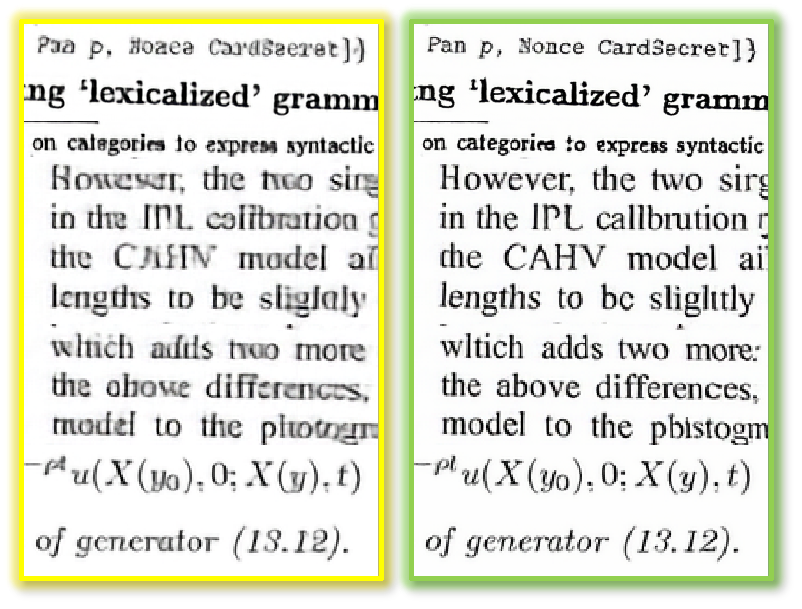}}
    \subfloat[\textbf{DocDiff}]{\includegraphics[width=0.14\linewidth, height=0.2\linewidth]{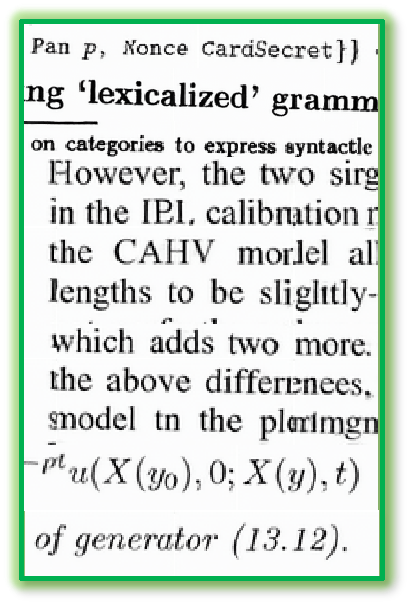}}
    \caption{Qualitative representative results of text line deblurring on the Document Deblurring Dataset \cite{hradivs2015convolutional}. }
    \label{fig:deblur}
\end{figure*}

\textbf{Watermark and Seal Removal:} 
Dense watermarks and variable seals greatly affect the readability and recognizability of covered characters. There is limited research on this issue in the document analysis community and a paucity of publicly available benchmark datasets. Thus, we synthesized paired datasets (document image with dense watermarks and seals and its corresponding clean version, see synthetic details in Synthetic Datasets section and Fig. \ref{fig:watermark_examples} in Appendix) using in-house data for training and testing. For the discussion of experimental results on the synthetic datasets, please refer to Section \ref{sec:ws} in Appendix.

We jointly train the CP and HRR modules by minimizing the $\mathcal{L}_{total}$ in Eq. \ref{loss:total} with $\beta_0=2$ and $\beta_1=0.5$. The total time steps $T$ are set to 100. We use random 128×128 crops with a batchsize of 64 during training, and evaluate DocDiff on both non-native (larger size patches or full-size images) and native (128×128 crop-predict-merge) resolutions and different sampling steps (5, 20, 50 and 100). For data augmentation, we perform random rotation and horizontal flip. The number of training iterations is one million.

\subsection{Evaluation Metrics}

 We employ the SOTA no-reference (NR) image quality assessment (IQA) methods, including MUSIQ \cite{Ke_2021_ICCV} that is sensitive to varying sizes and aspect ratios and MANIQA \cite{Yang_2022_CVPR} that won the NTIRE 2022 NR-IQA Challenge \cite{Gu_2022_CVPR}, as well as widely-used full-reference (FR) IQA methods including LPIPS \cite{Zhang_2018_CVPR} and DISTS \cite{9298952}, to evaluate the reconstruction quality of document images. We still compute PSNR and SSIM \cite{1284395} for completeness, although not the primary metrics.

For high-level binarization tasks, we evaluate methods on three evaluation metrics commonly used in competitions, including the FMeasure (FM), pseudo-FMeasure (p-FM) \cite{Ntirogiannis2013}, and PSNR. For removal tasks, we evaulate methods on four metrics: MANIQA \cite{Yang_2022_CVPR}, LPIPS \cite{Zhang_2018_CVPR}, PSNR and SSIM \cite{1284395}.

\subsection{Document Deblurring}
\subsubsection{Ablation Study}
We conduct the ablation experiments on the Document Deblurring Dataset \cite{hradivs2015convolutional} to verify the benefits of proposed components of DocDiff: High-Frequency Residual Refinement Module (HRR), Frequency Separation Training (FS). Additionally, we investigate the impact of resolutions (native or non-native), sampling method (stochastic or deterministic), and predicted target ($x_0$ or $\epsilon$) on the performance of the model. Table \ref{tab:abl} shows the quantitative results. (see qualitative results in Fig. \ref{img:abl} in Appendix.)

\textbf{HRR:} The HRR module effectively improves the perceptual quality by sharpening text edges. To further prove that the effectiveness of HRR is not solely due to an increase in parameters, we cascade an identical Unet structure (CR) behind CP to transform the model into a two-stage regression method. Experimental results show that while simply cascading more encoder-decoder layers improve PSNR and SSIM, the perceptual quality remain poor with blurred text edges. This reiterates the effectiveness of the HRR module.

\textbf{FS: } As shown in Table \ref{tab:abl}, training with frequency separation demonstrates an improvement in the perceptual quality and a reduction in distortion, thus achieving a better Perception-Distortion trade-off. This decoupled strategy can effectively enhance the capacity of HRR module to recover high-frequency information.

\textbf{Native or Non-native ? : } Performing the crop-predict-merge strategy at native resolution can significantly reduce the distortion. However, inferring at non-native resolution (full-size image) also yields competitive perceptual quality and distortion. In practice, this is a time-quality trade-off. For instance, at 300x300 resolution, using 128x128 crop-predict-merge inference requires processing 60\% more ineffective pixels compared to full-size inference.

\begin{table}[!t]
\caption{Results on the Document Deblurring Dataset \cite{hradivs2015convolutional}. The bold numbers represent the improvement of the HRR module over the regression-based baseline. Note that the weights of the HRR module come from DocDiff and no joint training with the baseline has been performed.}
\label{tab:HRR}
\resizebox{\columnwidth}{!}{%
\begin{tabular}{cclcccccc}
\hline
\multicolumn{3}{c}{\multirow{3}{*}{Method}} & \multicolumn{4}{c}{\textbf{Perceptual}} & \multicolumn{2}{c}{\textbf{Distortion}} \\ \cline{4-9} 
\multicolumn{3}{c}{} & \multicolumn{2}{c}{NR} & \multicolumn{2}{c}{FR} & \multicolumn{2}{c}{} \\ \cline{4-9} 
\multicolumn{3}{c}{} & MANIQA↑ & MUSIQ↑ & DISTS↓ & LPIPS↓ & PSNR↑ & SSIM↑ \\ \hline
\multicolumn{8}{c}{\textbf{\colorbox{pink!20}{Natural scenes}}} \\ 
\multicolumn{3}{c}{DeBlurGAN-v2 \cite{kupyn2019deblurgan}, ICCV2019} & 0.6967 & 50.25 & 0.1014 & 0.0991 & 20.76 & 0.8731 \\
\multicolumn{3}{c}{+HRR} & \textbf{0.7097} & \textbf{50.66} & \textbf{0.0952} & \textbf{0.0989} & 20.62 & \textbf{0.8736} \\
\multirow{2}{*}{\begin{tabular}[c]{@{}c@{}}Pec. of\\ Pages\end{tabular}} & \multicolumn{2}{c}{Better than GT} & 38.78\% & 48.83\% &  &  &  &  \\
 & \multicolumn{2}{c}{Better than DeBlurGAN-v2} & 68.53\% & 82.76\% &  &  &  &  \\ 
\multicolumn{3}{c}{MPRNet\cite{zamir2021multi}, CVPR2021} & 0.6675 & 47.52 & 0.1555 & 0.0900 & 21.27 & 0.8803 \\
\multicolumn{3}{c}{+HRR} & \textbf{0.6852} & \textbf{49.87} & \textbf{0.1384} & \textbf{0.0887} & 20.86 & 0.8768 \\
\multirow{2}{*}{\begin{tabular}[c]{@{}c@{}}Pec. of\\ Pages\end{tabular}} & \multicolumn{2}{c}{Better than GT} & 17.87\% & 40.12\% &  &  &  &  \\
 & \multicolumn{2}{c}{Better than MPRNet} & 57.04\% & 87.04\% &  &  &  &  \\ 
\multicolumn{3}{c}{HINet \cite{chen2021hinet}, CVPR2021} & 0.6836 & 47.59 & 0.1232 & 0.1163 & 24.15 & 0.9164 \\
\multicolumn{3}{c}{+HRR} & \textbf{0.7041} & \textbf{50.44} & \textbf{0.0963} & \textbf{0.0987} & 23.44 & 0.9158 \\
\multirow{2}{*}{\begin{tabular}[c]{@{}c@{}}Pec. of\\ Pages\end{tabular}} & \multicolumn{2}{c}{Better than GT} & 31.07\% & 46.67\% &  &  &  &  \\
 & \multicolumn{2}{c}{Better than HINet} & 56.91\% & 91.82\% &  &  &  &  \\ \hline
 \multicolumn{8}{c}{\colorbox{green!20}{\textbf{Document scenes}}} \\ 
 \multicolumn{3}{c}{DE-GAN \cite{souibgui2020gan}, TPAMI2020} & 0.6546 & 46.75 & 0.0968 & 0.0843 & 22.30 & 0.9155 \\
\multicolumn{3}{c}{+HRR} & \textbf{0.6973} & \textbf{50.36} & \textbf{0.0776} & \textbf{0.0696} & 21.21 & 0.9114 \\
\multirow{2}{*}{\begin{tabular}[c]{@{}c@{}}Pec. of\\ Pages\end{tabular}} & \multicolumn{2}{c}{Better than GT} & 23.48\% & 45.34\% &  &  &  &  \\
 & \multicolumn{2}{c}{Better than DE-GAN} & 79.89\% & 95.40\% &  &  &  &  \\ 
\multicolumn{3}{c}{DocEnTr \cite{souibgui2022docentr}, ICPR2022} & 0.5821 & 46.53 & 0.1802 & 0.2225 & 22.66 & 0.9130  \\
\multicolumn{3}{c}{+HRR} & \textbf{0.6637} & \textbf{51.84} & \textbf{0.1378} & \textbf{0.1653} & 21.65 & \textbf{0.9142}  \\
\multirow{2}{*}{\begin{tabular}[c]{@{}c@{}}Pec. of\\ Pages\end{tabular}} & \multicolumn{2}{c}{Better than GT} &5.57\% &68.57\%  &  &  &  &  \\
 & \multicolumn{2}{c}{Better than DocEnTr} &94.40\%  &93.71\%  &  &  &  &  \\ \hline
\multicolumn{3}{c}{GT} & 0.7207 & 51.03 & 0.0 & 0.0 & $\infty$ & 1.0 \\ \hline
\end{tabular}%
}
\end{table}

\textbf{Predict $x_0$ or $\epsilon$, Stochastic or Deterministic ? : } We employ the HRR module to predict the $\epsilon$ and perform original stochastic sampling \cite{ho2020denoising}. Except for NR-IQA, experimental results show inferior performance in FR-IQA, PSNR, and SSIM for this approach. On one hand, the short sampling step (100) restricts the capability of the noise prediction-based models. On the other hand, stochastic sampling results can lead to poor integration of generated characters with the given conditions, which leads to increased occurrence of substitution characters. This is also the reason for the high NQ-IQA score and low FQ-IQA score. These issues are addressed by combining the prediction of $x_0$ and deterministic sampling. Moreover, we notice that the approach based on predicting $x_0$ can effectively generate high-quality images with 5 sampling steps (see Tab. \ref{tab:deblur} and Tab. \ref{tab:stage} in Appendix), whereas the approach relying on predicting $\epsilon$ fails to achieve the same level of performance.
\subsubsection{Performance}
For a comprehensive comparison, we compare with SOTA document deblurring methods \cite{souibgui2020gan, souibgui2022docentr} as well as natural scene deblurring methods \cite{kupyn2019deblurgan, chen2021hinet, zamir2021multi, whang2022deblurring}. Table \ref{tab:deblur} shows quantitative results and Figure \ref{fig:deblur} shows qualitative results. DocDiff(Native) achieves the \textbf{best} MANIQA, DISTS, LPIPS, and SSIM metrics, while also achieving competitive PSNR and MUSIQ. Notably, we obtain the LPIPS of 0.0307, \textbf{a 66\% reduction} compared to MPRNet \cite{zamir2021multi} and \textbf{a 64\% reduction} compared to DE-GAN \cite{souibgui2020gan}. DocDiff uses only \textbf{one-fourth} of the parameters used by those two methods. We also compare DocDiff with SOTA diffusion-based deblurring method \cite{whang2022deblurring}, which predicts $\epsilon$ and samples stochastically. \cite{whang2022deblurring} can produce high-quality images (higher MUSIQ). However, its FR-IQA and distortion metrics are significantly worse than DocDiff's at the same sampling step. Moreover, DocDiff (Non-native) outperforms MPRNet\cite{zamir2021multi}, HINet\cite{chen2021hinet}, and two document-scene methods \cite{souibgui2020gan, souibgui2022docentr} in \textbf{all perceptual metrics} with only \textbf{5-step} sampling, while maintaining competitive distortion metrics. 

\begin{table}[!t]
\caption{Document deblurring results on the Document Deblurring Dataset \cite{hradivs2015convolutional}. DocDiff outperforms state-of-the-art deblurring regression methods for both natural and document scenes on all perceptual metrics, even at non-native resolutions, while maintaining competitive PSNR and SSIM scores. DocDiff-n means applying n-step sampling ($T$).}
\label{tab:deblur}
\resizebox{\columnwidth}{!}{%
\begin{tabular}{clccccccc}
\hline
\multicolumn{2}{c}{\multirow{3}{*}{Method}} & \multirow{3}{*}{Parameters} & \multicolumn{4}{c}{\textbf{Perceptual}} & \multicolumn{2}{c}{\textbf{Distortion}} \\ \cline{4-9} 
\multicolumn{2}{c}{} &  & \multicolumn{2}{c}{NR} & \multicolumn{2}{c}{FR} & \multicolumn{2}{c}{} \\ \cline{4-9} 
\multicolumn{2}{c}{} &  & MANIQA↑ & MUSIQ↑ & DISTS↓ & LPIPS↓ & PSNR↑ & SSIM↑ \\ \hline
\multicolumn{9}{c}{\textbf{\colorbox{pink!20}{Natural scenes}}} \\ 
\multicolumn{2}{c}{DeBlurGAN-v2 \cite{kupyn2019deblurgan}, ICCV2019} & 67M & 0.6967 & 50.25 & 0.1014 & 0.0991 & 20.76 & 0.8731 \\
\multicolumn{2}{c}{MPRNet\cite{zamir2021multi}, CVPR2021} & 34M & 0.6675 & 47.52 & 0.1555 & 0.0900 & 21.27 & 0.8803 \\
\multicolumn{2}{c}{HINet \cite{chen2021hinet}, CVPR2021} & 86M & 0.6836 & 47.59 & 0.1232 & 0.1163 & \textcolor{red}{24.15} & 0.9164 \\
\multicolumn{2}{c}{Whang et al. \cite{whang2022deblurring}, CVPR2022} & 33M & 0.6898 & \textcolor{red}{50.86} & 0.0830 & 0.0750 &19.89 &0.8742\\
\hline
\multicolumn{9}{c}{\colorbox{green!20}{\textbf{Document scenes}}} \\ 
\multicolumn{2}{c}{DE-GAN \cite{souibgui2020gan}, TPAMI2020} & 31M & 0.6546 & 46.75 & 0.0968 & 0.0843 & 22.30 & 0.9155 \\
\multicolumn{2}{c}{DocEnTr \cite{souibgui2022docentr}, ICPR2022} & 67M & 0.5821 & 46.53 & 0.1802 & 0.2225 & 22.66 & 0.9130 \\ \hline
\multicolumn{2}{c}{DocDiff (Non-native)-5} & 8.20M & 0.6873 & 47.92 & 0.0907 & 0.0582 & 22.17 & \textcolor{blue}{0.9223} \\ 
\multicolumn{2}{c}{DocDiff (Non-native)-100} & 8.20M & \textcolor{blue}{0.6971} & 50.31 & \textcolor{blue}{0.0636} & \textcolor{blue}{0.0474} & 20.46 & 0.9006 \\
\multicolumn{2}{c}{DocDiff (Native)-100} & 8.20M & \textcolor{red}{0.7174} & \textcolor{blue}{50.62} & \textcolor{red}{0.0611} & \textcolor{red}{0.0307} & \textcolor{blue}{23.28} & \textcolor{red}{0.9505} \\ \hline
\multicolumn{2}{c}{GT} & - & 0.7207 & 51.03 & 0.0 & 0.0 & $\infty$ & 1.0 \\ \hline
\end{tabular}%
}
\end{table}

To validate the universality of the HRR module, we refine the output of baselines \cite{kupyn2019deblurgan, chen2021hinet, zamir2021multi, souibgui2020gan, souibgui2022docentr} directly using the pre-trained HRR module. As shown in Tab. \ref{tab:HRR}, after refinement, \textbf{all perceptual metrics} are improved. Specifically, the DISTS of HINet \cite{chen2021hinet} decrease by \textbf{22\%}, and the MANIQA of DocEnTr \cite{souibgui2022docentr} increase by \textbf{14\%}. We calculate the percentage of samples in which NR-IQA show better performance compared to the GT and baselines after refinement. On average, in terms of MUSIQ, \textbf{90\%} of the samples show improvement over baselines, and \textbf{50\%} over GT. As shown in Figs. \ref{fig:kaipian} and \ref{fig:deblur}, DocDiff provides the most clear and accurate capability of character pixel restoration. After refined by the HRR module, the text edges of the baselines become sharp, but wrong characters still exist.

\subsection{Document Denoising and Binarization}
\begin{table}[!t]
\caption{Results of document binarization on H-DIBCO 2018 \cite{Pratikakis2018} and DIBCO 2019 \cite{Pratikakis2019}. Best values are \textbf{bold}. \label{tab:binarization}}
\centering
\resizebox{\linewidth}{!}{%
\begin{tabular}{cccccccccc}
\hline
\multirow{2}{*}{Method}  & \multirow{2}{*}{Parameters} & \multicolumn{3}{c}{H-DIBCO'18}                                   & \multicolumn{3}{c}{DIBCO'19} \\ \cline{3-8} 
                         &                             & FM↑             & p-FM↑           & PSNR↑                     & FM↑                 & p-FM↑              & PSNR↑                          \\ \hline
Otsu \cite{Otsu1979}                    & -                           & 51.45          & 53.05          & 9.74                    & 47.83              & 45.59             & 9.08                     \\
Sauvola \cite{Sauvola2000}, PR2000          & -                           & 67.81          & 74.08          & 13.78                 & 51.73              & 55.15             & 13.72                        \\
Howe \cite{Howe2013}, IJDAR2013          & -                           & 80.84          & 82.85          & 16.67                & 48.20              & 48.37             & 11.38                        \\
Jia et al. \cite{Jia2018}, PR2018       & -                           & 76.05          & 80.36          & 16.90                    & 55.87              & 56.28             & 11.34                         \\
Kligler et al. \cite{kligler2018document}, CVPR2018 & -                           & 66.84          & 68.32          & 15.99           & 53.49              & 53.34             & 11.23                   \\
1st rank of contest      & -                           & 88.34          & 90.24          & \textbf{19.11}           & 72.88              & 72.15             & 14.48                        \\
cGANs \cite{Zhao2019}, PR2019            & 103M                     & 87.73          & 90.60          & 18.37             & 62.33              & 62.89             & 12.43                         \\
DE-GAN \cite{souibgui2020gan}, TPAMI2020        & 31M                      & 77.59          & 85.74          & 16.16                 & 55.98              & 53.44             & 12.29                     \\
$\mathrm{D}^2$ BFormer \cite{yang2023novel}, IF2023        & 194M                     & \textbf{88.84} & \textbf{93.42} & 18.91           & 67.63              & 66.69             & 15.05            \\
DocDiff                  & 8.20M                       & 88.11          & 90.43          & 17.92                    & \textbf{73.38}     & \textbf{75.12}    & \textbf{15.14}               \\ \hline
\end{tabular}}
\end{table}
\begin{figure}[!t]
    \centering
    \captionsetup[subfloat]{labelsep=none, format=plain, labelformat=empty}
    \subfloat[\textbf{Input}]{\includegraphics[width=0.33\linewidth, height=0.33\linewidth]{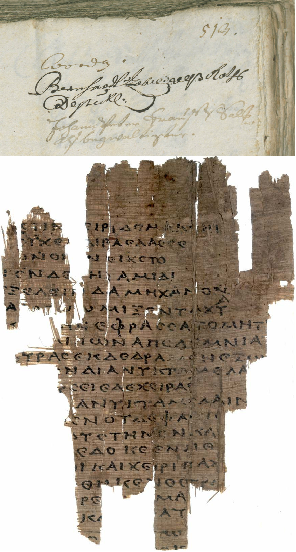}}
    \subfloat[\textbf{Ground-truth}]{\includegraphics[width=0.33\linewidth, height=0.33\linewidth]{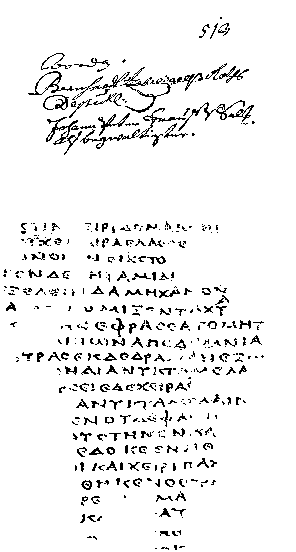}}
    \subfloat[\textbf{Sauvola \cite{Sauvola2000}}]{\includegraphics[width=0.33\linewidth, height=0.33\linewidth]{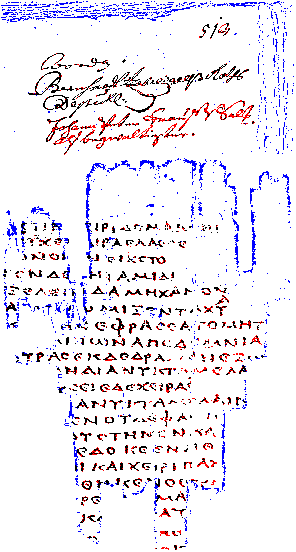}}
    \quad
    \subfloat[\textbf{Howe \cite{Howe2013}}]{\includegraphics[width=0.33\linewidth, height=0.33\linewidth]{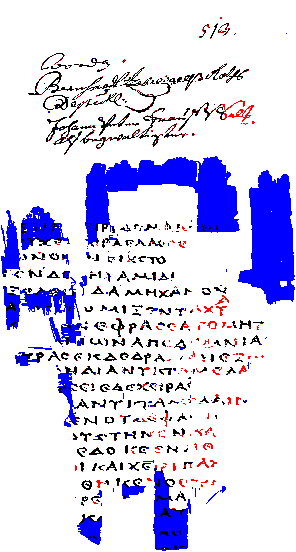}}
    \subfloat[\textbf{Kligler \cite{kligler2018document}}]{\includegraphics[width=0.33\linewidth, height=0.33\linewidth]{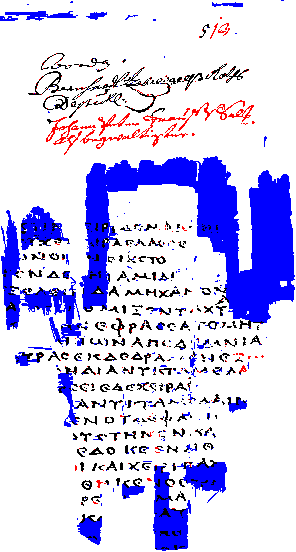}}
    \subfloat[\textbf{Jia \cite{Jia2018}}]{\includegraphics[width=0.33\linewidth, height=0.33\linewidth]{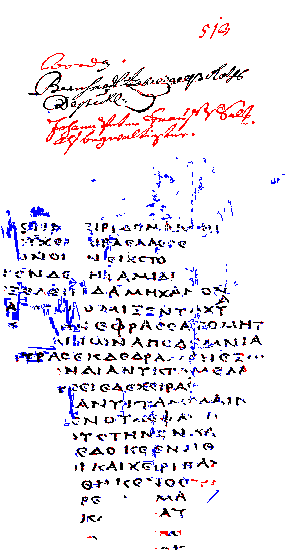}}
    \quad
    \subfloat[\textbf{cGANs \cite{Zhao2019}}]{\includegraphics[width=0.33\linewidth, height=0.33\linewidth]{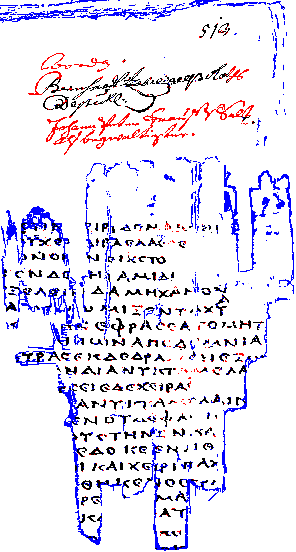}}
    \subfloat[\textbf{DE-GAN \cite{souibgui2020gan}}]{\includegraphics[width=0.33\linewidth, height=0.33\linewidth]{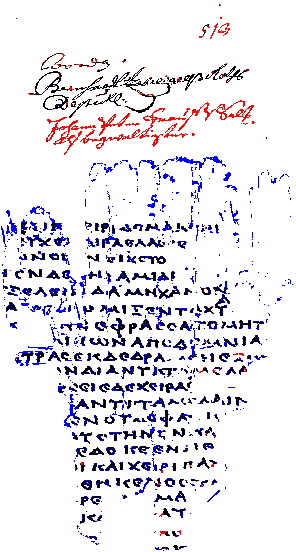}}
    \subfloat[\textbf{DocDiff}]{\includegraphics[width=0.33\linewidth, height=0.33\linewidth]{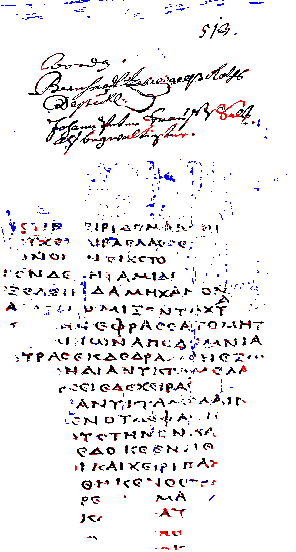}}
    \caption{Binarization results of two example images on H-DIBCO 2018 \cite{Pratikakis2018} and DIBCO 2019 \cite{Pratikakis2019}. Text pixels classified as background are highlighted in \textcolor{red}{red}, whereas background pixels classified as text are highlighted in \textcolor{blue}{blue}.}
    \label{fig:DIBCO}
\end{figure}
We compare with threshold-based methods \cite{Otsu1979, Sauvola2000, Howe2013, kligler2018document, Jia2018} and SOTA methods \cite{Zhao2019, souibgui2020gan, yang2023novel}. Quantitative and qualitative results are shown in Tab. \ref{tab:binarization} and Fig. \ref{fig:DIBCO}, respectively. On the H-DIBCO'18, our method do not achieve SOTA performance. However, our F-Measure is 10.52\% higher than that of DE-GAN \cite{souibgui2020gan}, and we also have competitive performance in terms of p-FM and PSNR. 
Due to the absence of papyri material in training sets and the presence of a large amount of newly introduced noise in DIBCO'19, this dataset poses a great challenge. DocDiff outperforms existing methods on DIBCO'19 with an F-Measure improvement of 5.75\% over $\mathrm{D}^2$ BFormer \cite{yang2023novel} and 11.05\% over cGANs \cite{Zhao2019}. Remarkably, this is achieved using significantly fewer parameters, with only 1/23 and 1/12 the parameter count of $\mathrm{D}^2$ BFormer \cite{yang2023novel} and cGANs \cite{Zhao2019}, respectively.

\begin{table}[!t]
\caption{Average runtimes (seconds/megapixel) of different methods.}
\label{tab:time}
\resizebox{\linewidth}{!}{%
\begin{tabular}{ccccc}
\hline
\textbf{Method} & MPRNet \cite{zamir2021multi} & DE-GAN \cite{souibgui2020gan} & DocDiff(Non-native)-5 & DocDiff(Non-native)-100 \\ \hline
\textbf{Runtime} & 0.57 & 0.82 & 0.33 & 5.69 \\ \hline
\end{tabular}%
}
\end{table}

Diffusion models are notorious for their time complexity of inference. DocDiff (Non-native)-5 achieve competitive performance in removing blur and watermarks. We compare the time complexity of our methods with MPRNet \cite{zamir2021multi} and DE-GAN \cite{souibgui2020gan} under the same hardware environment. However, due to different frameworks being used (Tensorflow for DE-GAN \cite{souibgui2020gan} while PyTorch for MPRNet \cite{zamir2021multi} and our methods), the speed comparison is only approximate. The results are shown in Tab. \ref{tab:time}. DocDiff (Non-native)-5 offers great efficiency and perfomance, making it an ideal choice for various document image enhancement tasks.

\subsection{OCR Evaluation}
\begin{figure}
    \centering
    \includegraphics[width=\linewidth]{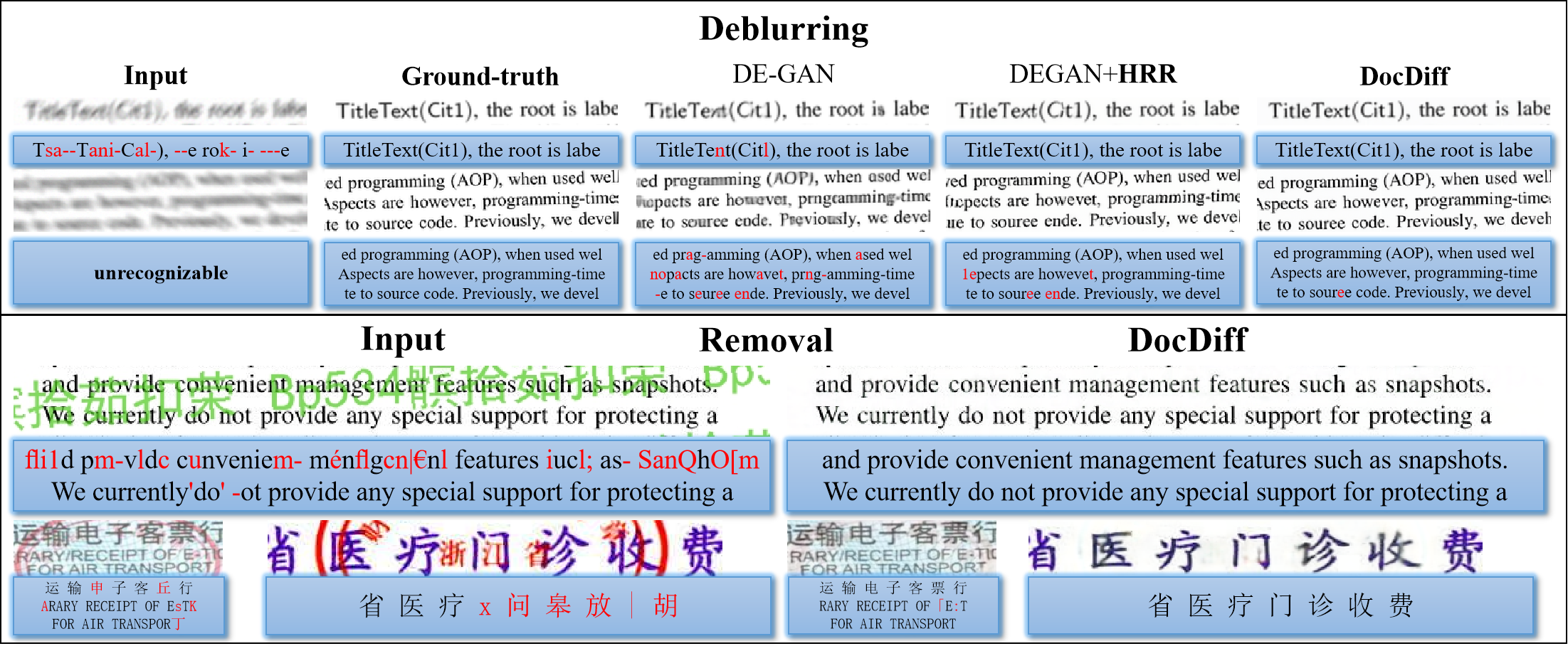}
    \caption{Qualitative results for Tesseract recognition of various text lines. The \textcolor{red}{red characters} indicate recognition errors or missed characters. Best viewed with zoom-in.}
    \label{fig:ocr}
\end{figure}

we compare the OCR performance on degraded and enhanced documents using a set of 50 text patches. This set includes 30 blurred patches from the Document Deblurring Dataset \cite{hradivs2015convolutional}, 10 patches with watermarks, and 10 patches with seals. We use Tesseract OCR to recognize those patches. Highly blurred images are barely recognizable. After applying DE-GAN \cite{souibgui2020gan}, DE-GAN\cite{souibgui2020gan}+HRR, and DocDiff for deblurring, the character error rates of the enhanced versions are reduced to 13.7\%, 7.6\%, and 4.4\%, respectively. For the removal task, the character error rates of the original images and the ones enhanced with DocDiff are 28.7\% and 1.8\%, respectively. The recognition results on some patches are shown in Fig. \ref{fig:ocr}.

\section{Conclusions}
In this paper, we propose a novel and unified framework, DocDiff, for various document enhancement tasks. DocDiff significantly improves the perceptual quality of reconstructed document images by utilizing a residual prediction-based conditional diffusion model. For the deblurring task, our proposed HRR module is ready-to-use, which effectively sharpens the text edges generated by regression methods \cite{kupyn2019deblurgan, chen2021hinet, zamir2021multi, souibgui2020gan, souibgui2022docentr} to enhance the readability and recognizability of the text. Compared to non-diffusion-based methods, DocDiff achieves competitive performance with only 5 steps of sampling, and is lightweight, which greatly optimizes its inference time complexity. We believe that DocDiff establishes a strong benchmark for future work.

% There are several potential solutions to overcome the limitations of our work. During the experiment, we observe that although DocDiff is able to generate sharp text edges in most cases, there are still instances where characters appear erroneous or distorted. To address this issue, one approach is to utilize the text prior \cite{zhu2022improving} in order to incorporate additional semantic information. Currently, there is a scarcity of paired large-scale document enhancement benchmark datasets in real-world scenarios. This leads to a lack of generalizability in practical applications. One approach to address this issue is to utilize DocDiff for assisted annotation. For instance, in the case of seal removal, humans can annotate on the results generated by DocDiff, thereby substantially reducing labelling complexity. Through multiple rounds of iteration and fine-tuning, the dataset can be expanded and the performance of the model can be improved.

\bibliographystyle{ACM-Reference-Format}
\bibliography{sample-base}

\appendix

\section{Degraded Documents}

\begin{figure}[!htbp]
    \centering
    \subfloat[Various noises (Broken holes, Bleed-through, Smear, Faint ink)]{\includegraphics[width=\linewidth]{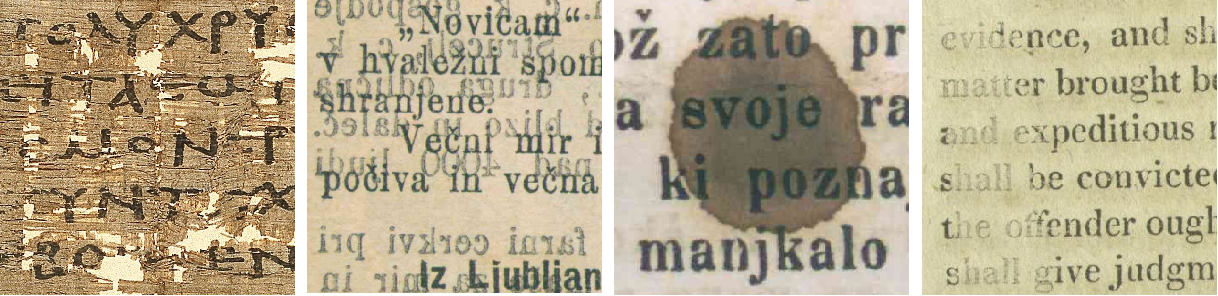}}
    \quad
    \subfloat[Blur]{\includegraphics[width=0.5\linewidth]{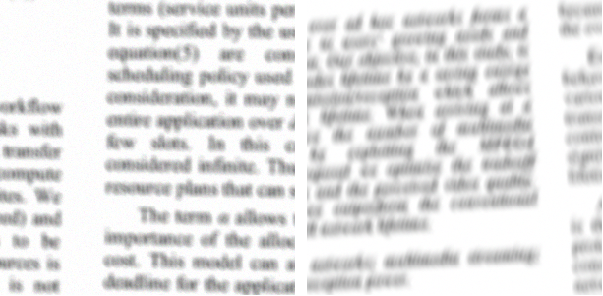}}
    \subfloat[Watermark]{\includegraphics[width=0.25\linewidth]{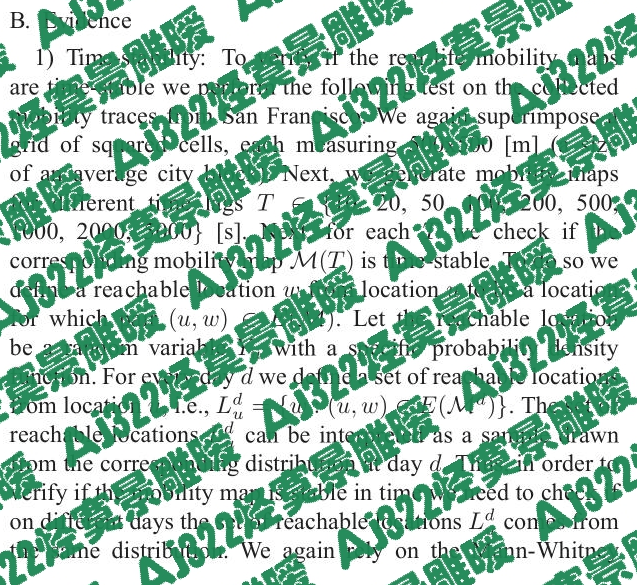}}
    \subfloat[Seal]{\includegraphics[width=0.25\linewidth]{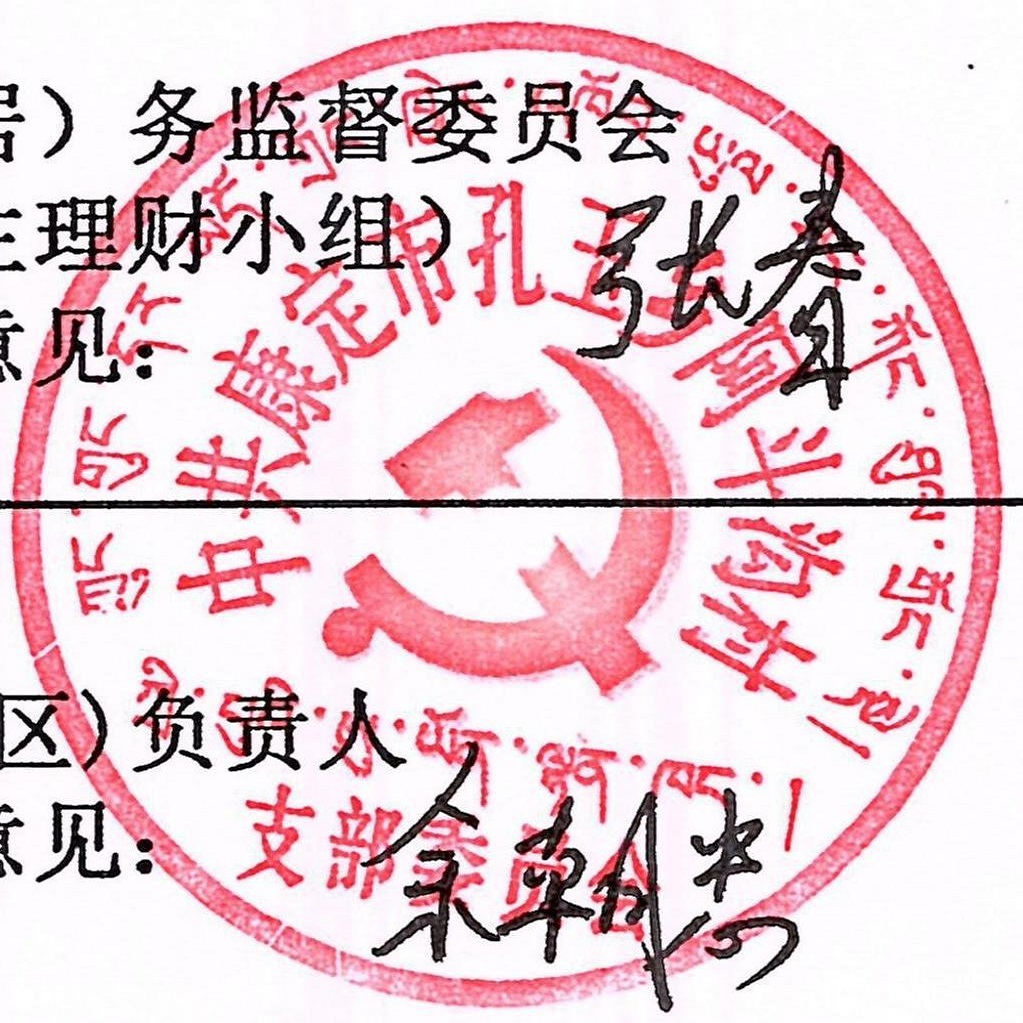}}
    \caption{Examples of degraded document images.}
    \label{fig:noise}
\end{figure}

Figure \ref{fig:noise} illustrates the three types of degraded documents that our work aims to enhance: documents with fragmented noise, blurry documents, and documents with seals and dense watermarks. Regarding the removal of seals, we mainly focus on red seals in the context of Chinese documents.

\section{Synthetic Datasets}
\begin{figure}[!htbp]
    \centering
    \subfloat{\includegraphics[width=0.20\linewidth, height=0.3\linewidth]{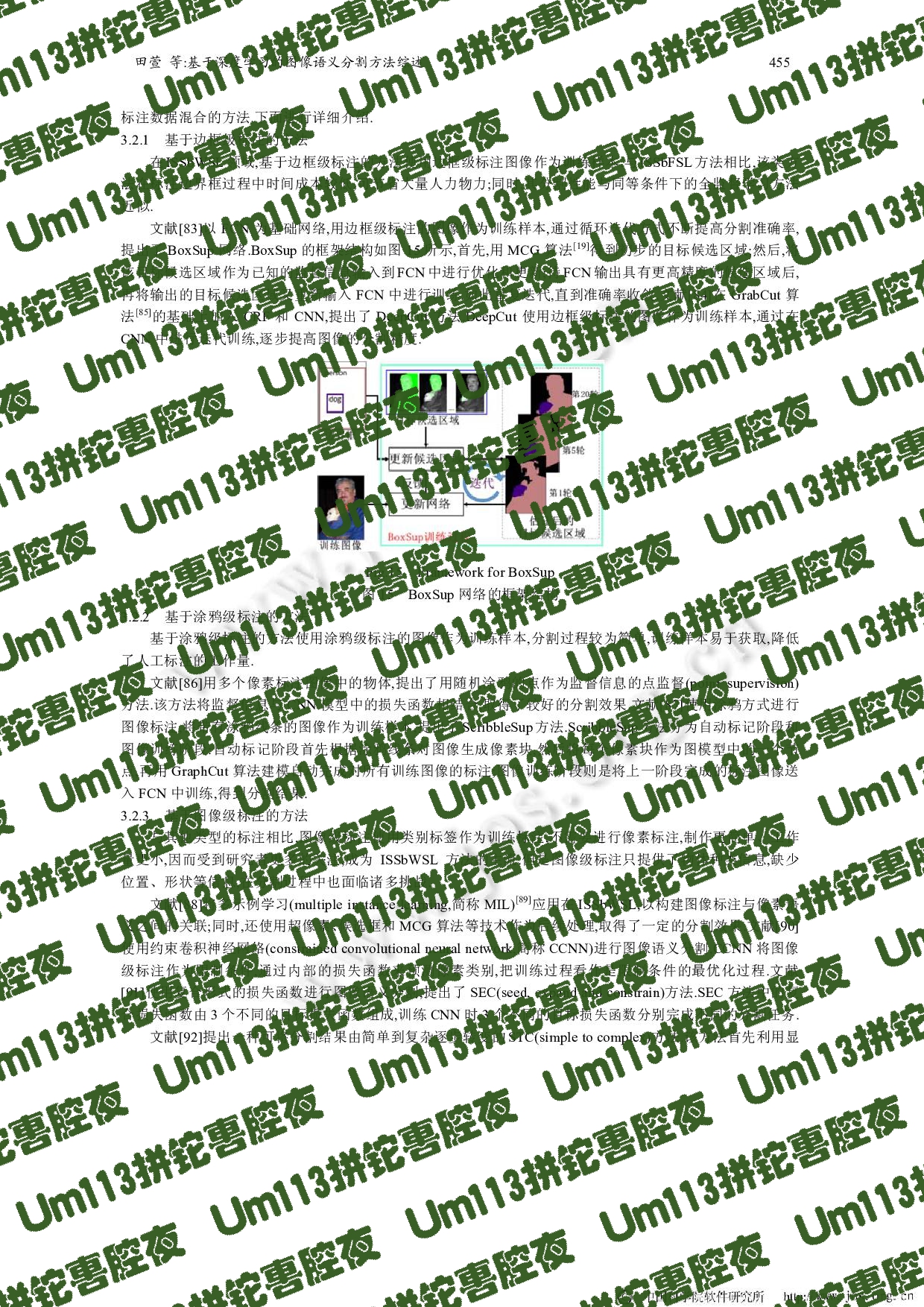}}
    \subfloat{\includegraphics[width=0.20\linewidth, height=0.3\linewidth]{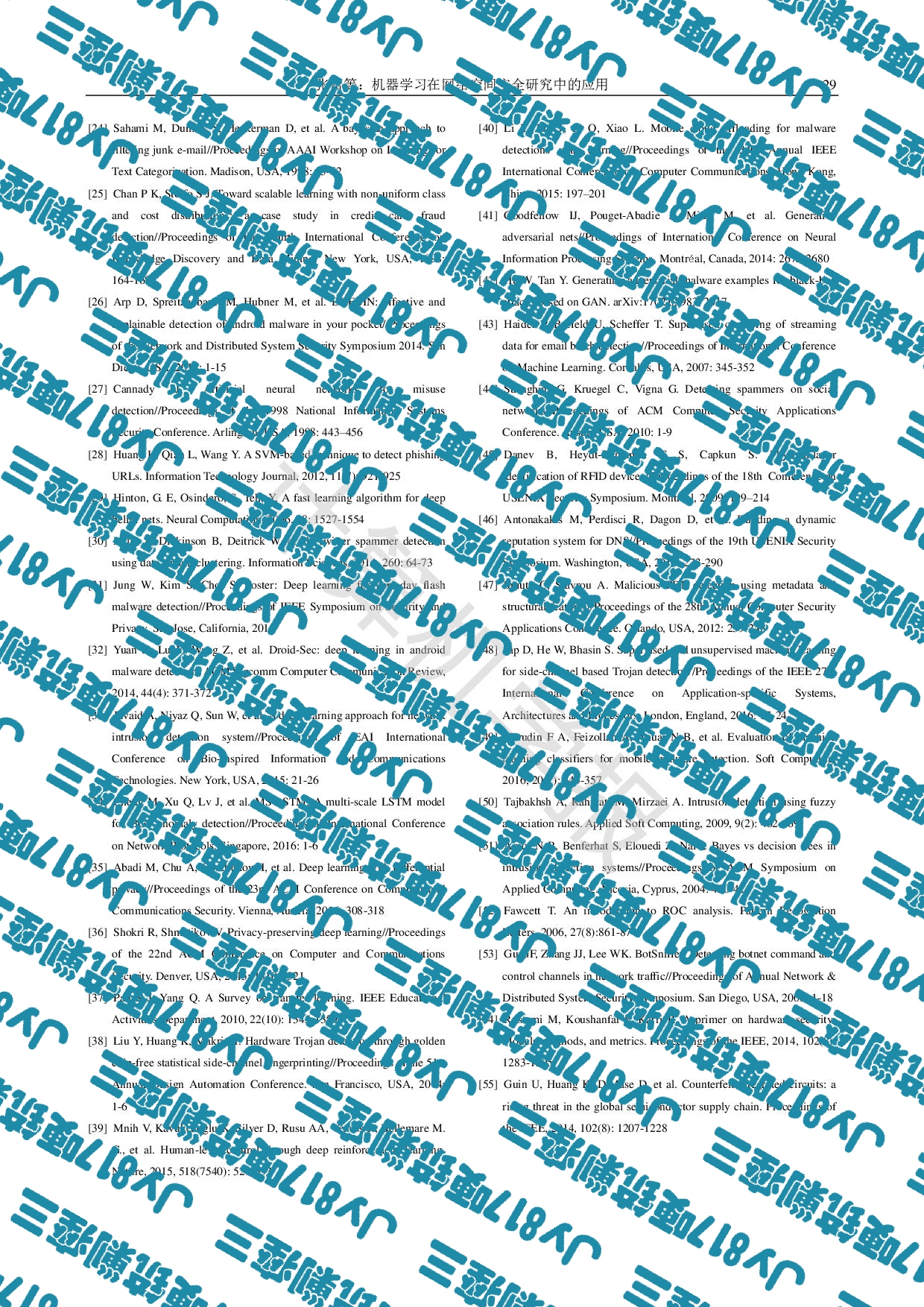}}
    \subfloat{\includegraphics[width=0.3\linewidth, height=0.3\linewidth]{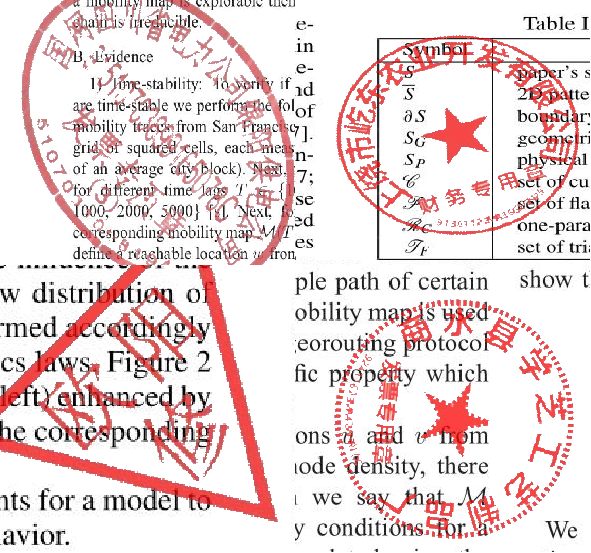}}
    \subfloat{\includegraphics[width=0.3\linewidth, height=0.3\linewidth]{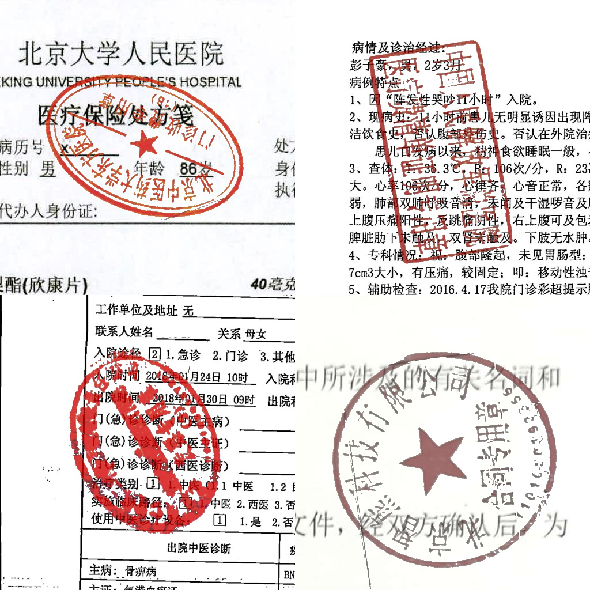}}
    \caption{Examples of document images synthesized with dense watermarks and real seals.}
    \label{fig:watermark_examples}
\end{figure}

Figure \ref{fig:watermark_examples} shows some examples from our synthetic datasets.
The synthesized dense watermarks feature randomized text (including both Chinese and English characters and numbers), font, size, color, spacing, position, and angle. The opacity of the watermarks is randomly sampled between 0.7 to 0.95. We utilized our unified seal-segmentation method to extract the mask of the seals in real Chinese document scenes. These seals mostly come from our internal documents, with a small portion coming from the ICDAR 2023 Competition on Reading the Seal Title. Afterwards, we fused the seal masks into the background images in the same manner. In our developed datasets, documents covered with watermarks have a resolution of 1754×1240 (with 3000 512×512 patches allocated for training and 100 full images for testing), while those covered with seals have a resolution of 512×512 (with 3000 for training and 500 used for testing). Note that the watermarks, seals, and background images in the training sets and testing sets are independent.

\section{Watermark and Seal Removal}
\label{sec:ws}
\begin{table}[!htbp]
\caption{Watermark and seal removal results on our developed datasets.}
\label{tab:ws_removal}
\resizebox{\linewidth}{!}{%
\begin{tabular}{clcccccccc}
\hline
\multicolumn{2}{c}{\multirow{3}{*}{Method}} & \multicolumn{4}{c}{Watermark Removal} & \multicolumn{4}{c}{Seal Removal} \\ \cline{3-10} 
\multicolumn{2}{c}{} & \multicolumn{2}{c}{Perceptual} & \multicolumn{2}{c}{Distortion} & \multicolumn{2}{c}{Perceptual} & \multicolumn{2}{c}{Distortion} \\
\multicolumn{2}{c}{} & MANIQA↑ & LPIPS↓ & PSNR↑ & SSIM↑ & MANIQA↑ & LPIPS↓ & PSNR↑ & SSIM↑ \\ \hline
\multicolumn{2}{c}{MPRNet \cite{zamir2021multi}} & 0.5253 & 0.0806 & \textbf{33.36} & 0.9497 & \textbf{0.5133} & \textbf{0.0913} & \textbf{32.76} & \textbf{0.9397} \\
\multicolumn{2}{c}{DE-GAN \cite{souibgui2020gan}} & 0.5190 & 0.1167 & 24.11 & 0.9106 & 0.4742 & 0.4115 & 19.62 & 0.7084 \\
\multicolumn{2}{c}{DocDiff(Non-native)-5} & 0.5263 & 0.0680 & 30.91 & \textbf{0.9637} & 0.5018 & 0.1162 & 31.07 & 0.9292 \\
\multicolumn{2}{c}{DocDiff(Non-native)-100} & \textbf{0.5267} & \textbf{0.0577} & 30.36 & 0.9576 & 0.5031 & 0.1152 & 30.67 & 0.9225 \\ \hline
\multicolumn{2}{c}{Ground Truth} & 0.5367 & 0.0 & $\infty$ & 1.0 & 0.5525 & 0.0 & $\infty$ & 1.0 \\ \hline
\end{tabular}%
}
\end{table}
\begin{figure}[!htbp]
    \centering
    \captionsetup[subfloat]{labelsep=none, format=plain, labelformat=empty}
    \subfloat[\textbf{Input}]{\includegraphics[width=0.2\linewidth, height=0.6\linewidth]{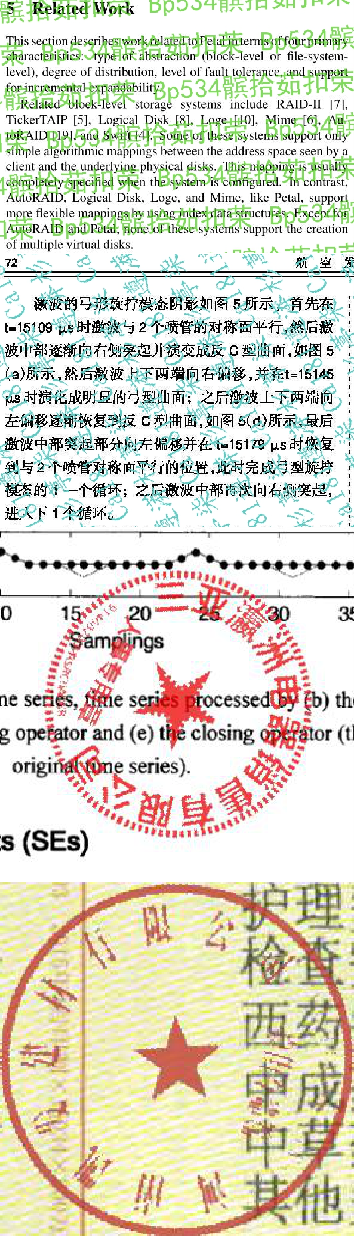}}
    \subfloat[\textbf{Ground-truth}]{\includegraphics[width=0.2\linewidth, height=0.6\linewidth]{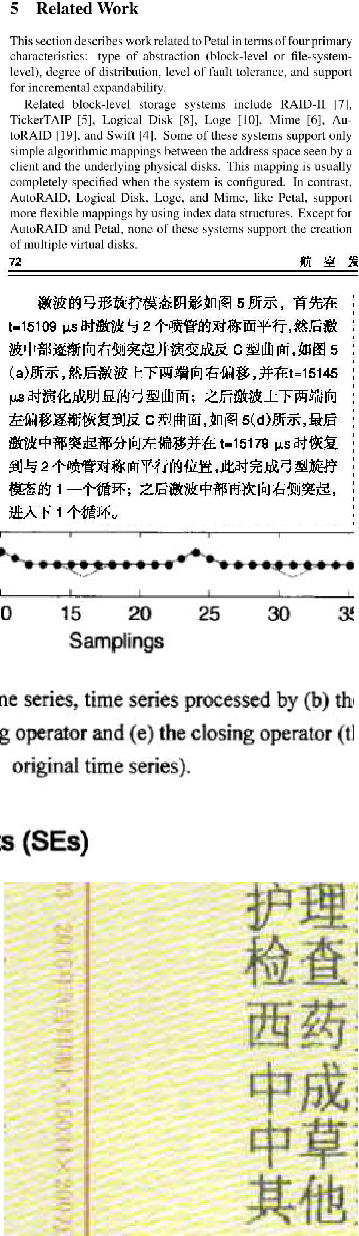}}
    \subfloat[\textbf{DE-GAN \cite{souibgui2020gan}}]{\includegraphics[width=0.2\linewidth, height=0.6\linewidth]{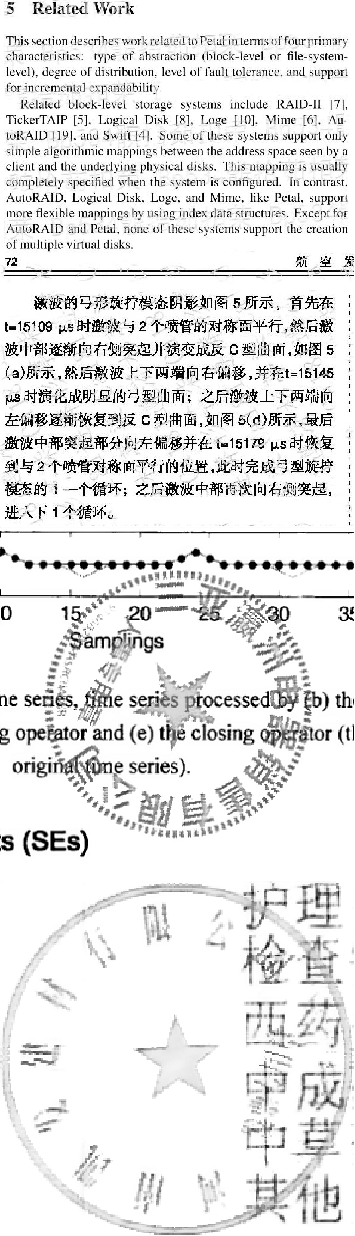}}
    \subfloat[\textbf{MPRNet \cite{zamir2021multi}}]{\includegraphics[width=0.2\linewidth, height=0.6\linewidth]{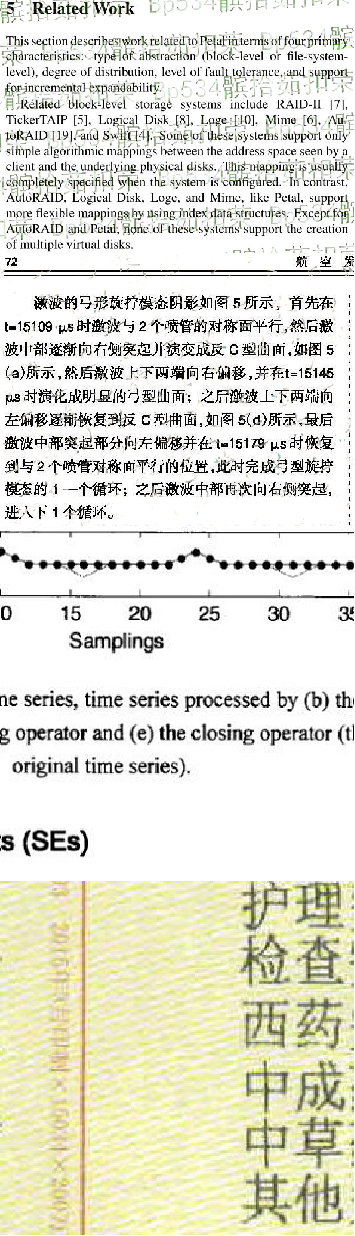}}
    \subfloat[\textbf{DocDiff-5}]{\includegraphics[width=0.2\linewidth, height=0.6\linewidth]{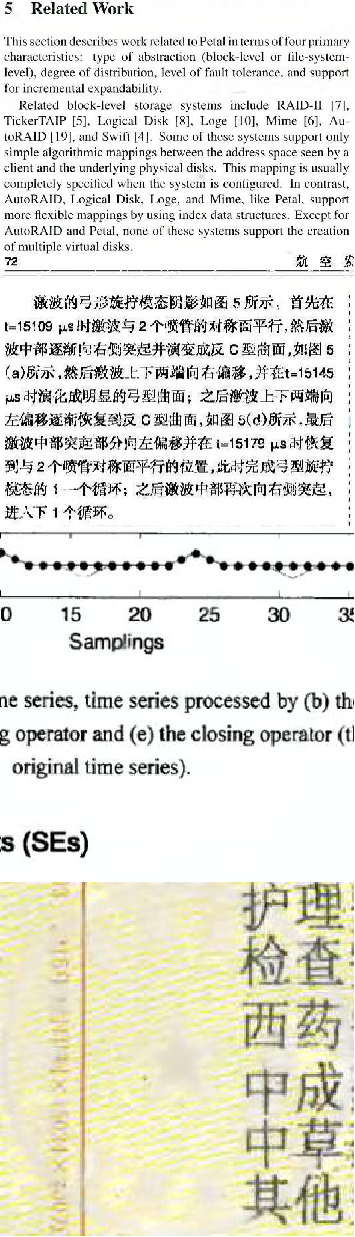}}
    \caption{Qualitative watermark and seal removal results on synthetic datasets. While DocDiff is effective at removing watermarks and seals, MPRNet \cite{zamir2021multi} displays better performance in restoring background on invoices.}
    \label{fig:wm_seal}
\end{figure}
\begin{figure}[!htbp]
    \centering
    \captionsetup[subfloat]{labelsep=none, format=plain, labelformat=empty}
    \subfloat[\textbf{Input}]{\includegraphics[width=0.2\linewidth, height=0.6\linewidth]{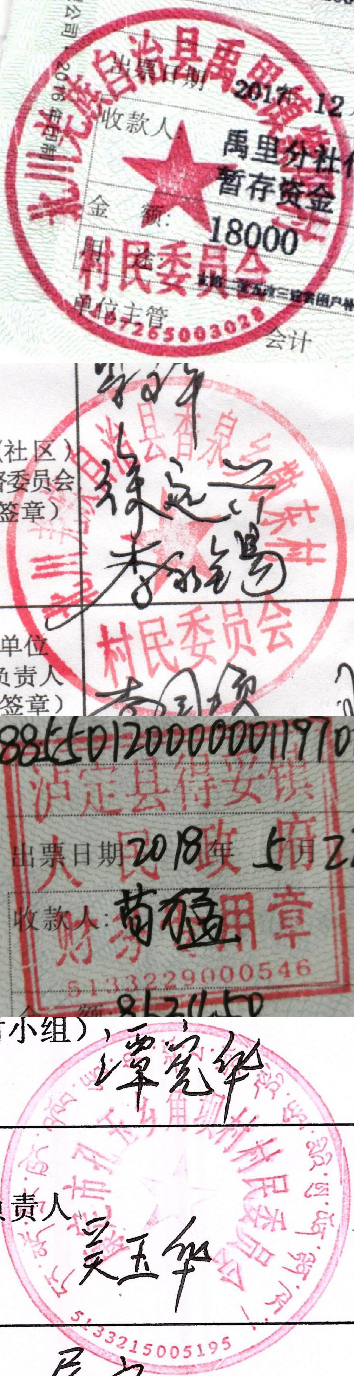}}
    \subfloat[\textbf{DE-GAN \cite{souibgui2020gan}}]{\includegraphics[width=0.2\linewidth, height=0.6\linewidth]{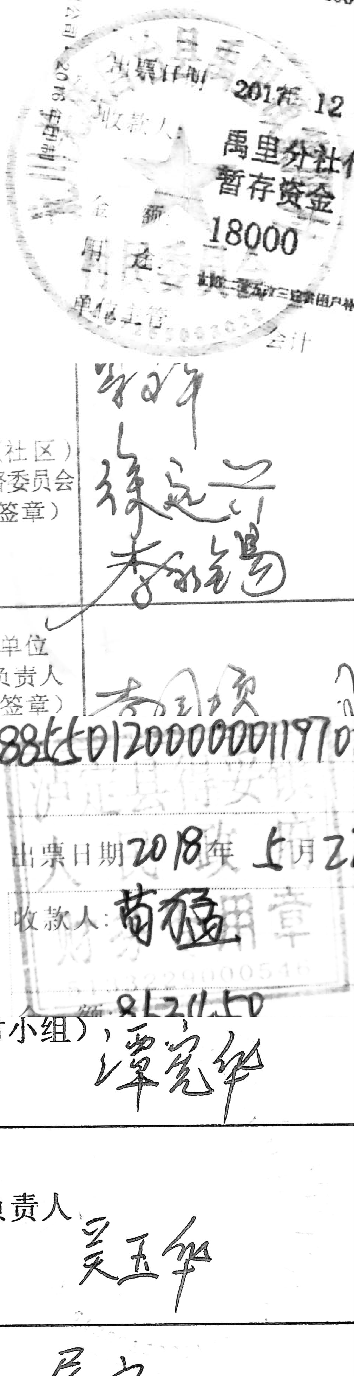}}
    \subfloat[\textbf{MPRNet \cite{zamir2021multi}}]{\includegraphics[width=0.2\linewidth, height=0.6\linewidth]{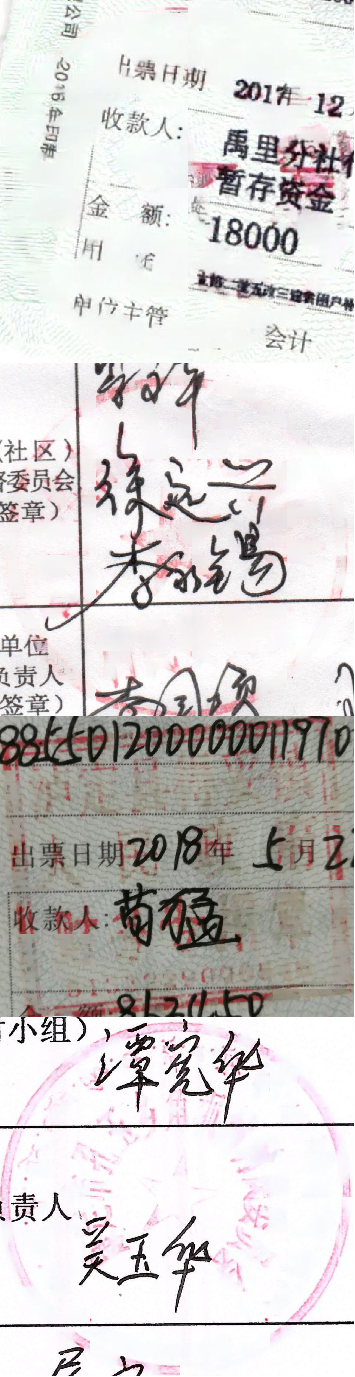}}
    \subfloat[\textbf{DocDiff-5}]{\includegraphics[width=0.2\linewidth, height=0.6\linewidth]{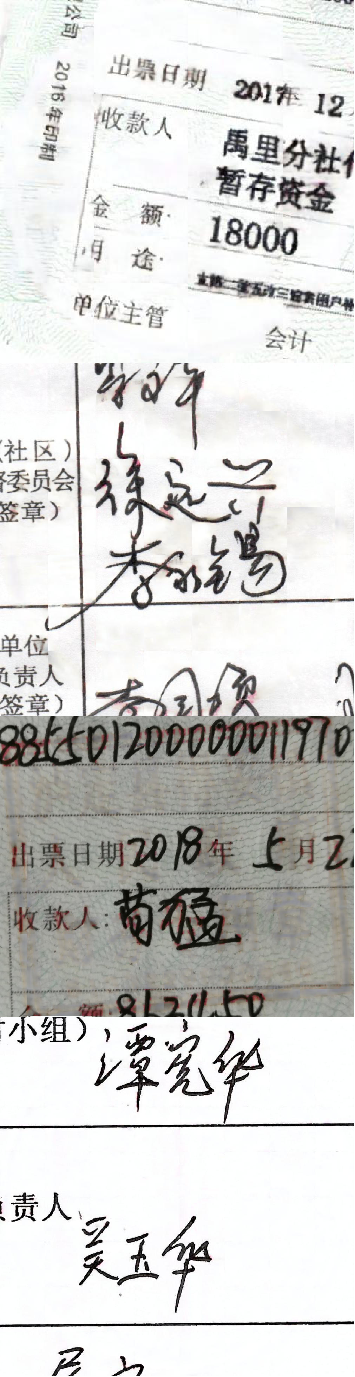}}
    \caption{Qualitative seal removal results in real Chinese invoice and document scenarios. Our proposed DocDiff model exhibits superior generalization ability and is resilient to noise.}
    \label{fig:seals}
\end{figure}

Quantitative and qualitative results are shown in Tab. \ref{tab:ws_removal} and Fig. \ref{fig:wm_seal}, respectively.
For watermark removal, DocDiff (Non-native)-5 exhibits competitive performance as compared to MPRNet \cite{zamir2021multi}. For seal removal, MPRNet \cite{zamir2021multi} perform better on synthetic datasets due to its well-designed feature extraction module that can effectively restore diverse invoice backgrounds. However, in real-world Chinese invoice and document scenarios, DocDiff demonstrate better generalization ability, as shown in Fig. \ref{fig:seals}. DE-GAN \cite{souibgui2020gan} is designed to take grayscale images as input and output, hence its performance on multi-color removal is bad.

\section{Metrics Description}

\begin{figure}[!htbp]
    \centering
    \includegraphics[width=\linewidth]{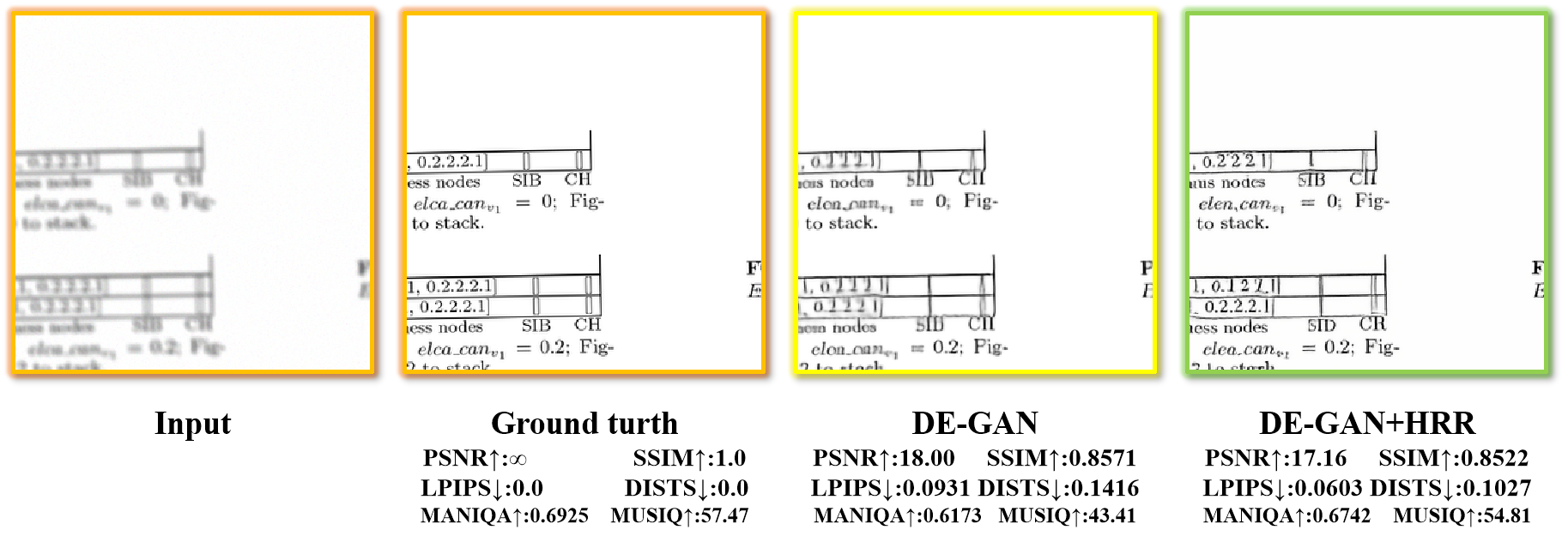}
    \caption{Comparison of different metrics. Best viewed with zoom-in.}
    \label{fig:metrics}
\end{figure}

As shown in Fig. \ref{fig:metrics}, DE-GAN generates images with higher PSNR and SSIM scores but the character pixel-level edges are notably blurred, causing difficulty in reading for humans and recognition for OCR systems. With the enhancement of the HRR module, the character edges become much sharper and easier to recognize. The improvement in multiple perceptual metrics aligns with human perceptual quality.

\section{More details in Ablation Study}

\begin{figure}[!htbp]
\centering
\captionsetup[subfloat]{labelsep=none, format=plain, labelformat=empty}
    \subfloat[\textbf{Input}]{\includegraphics[width=0.25\linewidth]{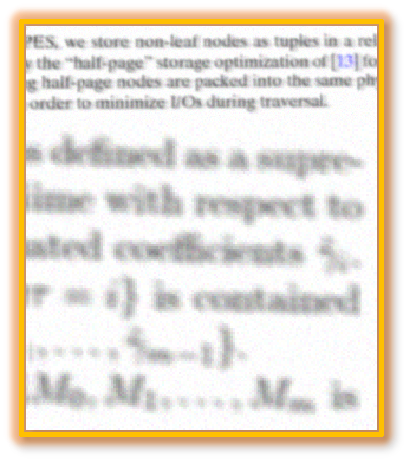}}
    \subfloat[\textbf{Ground-truth}]{\includegraphics[width=0.25\linewidth]{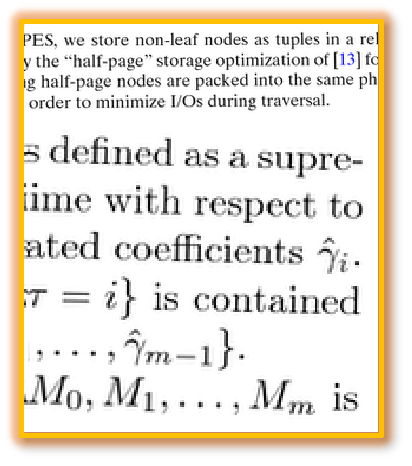}}
    \subfloat[\textbf{CP}]{\includegraphics[width=0.25\linewidth]{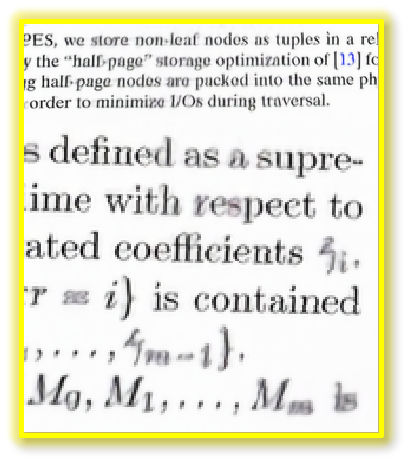}}
    \subfloat[\textbf{CP+CR}]{\includegraphics[width=0.25\linewidth]{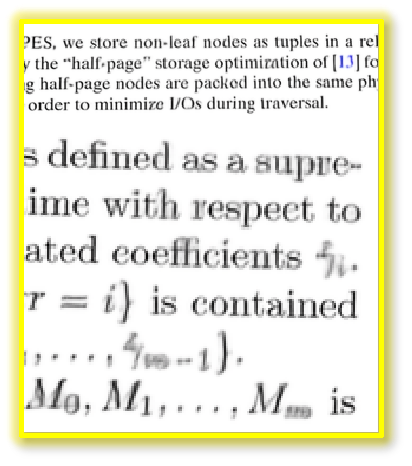}}
    \quad
    \subfloat[\textbf{CP+HRR (Stochastic)}]{\includegraphics[width=0.25\linewidth]{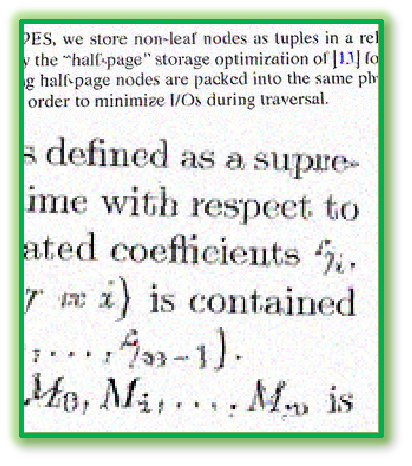}}
    \subfloat[\textbf{CP+HRR (w/o FS)}]{\includegraphics[width=0.25\linewidth]{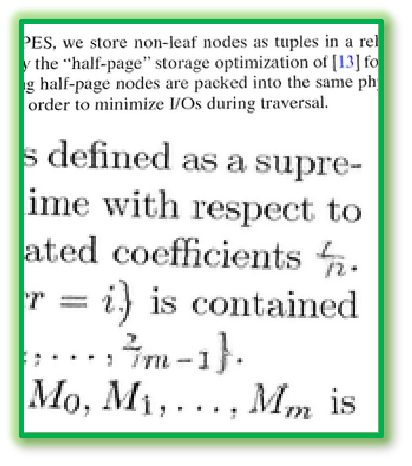}}
    \subfloat[\textbf{CP+HRR (w FS)}]{\includegraphics[width=0.25\linewidth]{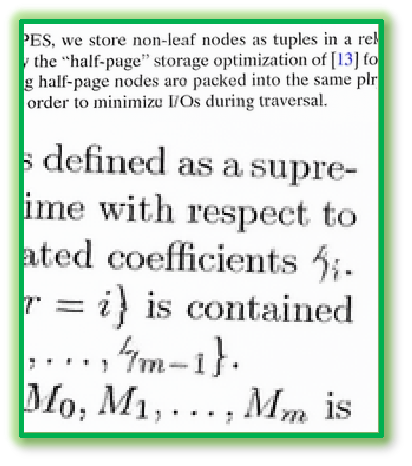}}
    \subfloat[\textbf{CP+HRR (Native)}]{\includegraphics[width=0.25\linewidth]{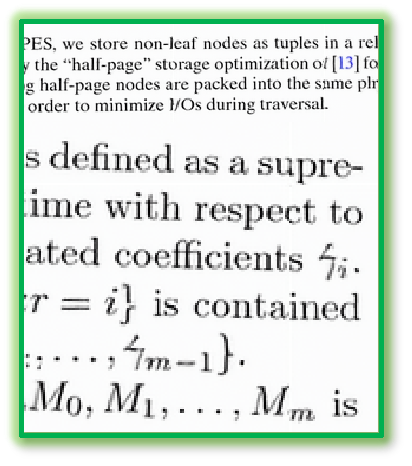}}
\centering
\caption{Qualitative ablation study results on the Document Deblurring Dataset. (CP: Coarse Predictor, CR: Cascade Refinement Module, HRR: High-Frequency Residual Refinement Module, FS: Frequency Separation Training.}
\label{img:abl}
\centering
\end{figure}

Figure~\ref{img:abl} shows qualitative results in ablation study. We can get the following useful conclusions:
\begin{itemize}
    \setlength{\itemsep}{0pt}
    \setlength{\parsep}{0pt}
    \setlength{\parskip}{0pt}
  \item Adding more encoder-decoder layers to optimize pixel loss in a cascade manner may not necessarily improve edge sharpening and legibility.
  \item Training with frequency separation can better restore text edges.
  \item Predicting added noise $\epsilon$ and applying short-step stochastic sampling can result in noisy sampled images with less sharp text edges.
  \item Inference at native resolution provides the best performance.
\end{itemize}

\section{Discussion about sampling steps}

\begin{figure}[!h]
    \centering
    \captionsetup[subfloat]{labelsep=none, format=plain, labelformat=empty}
    \subfloat[\textbf{Input}]{\includegraphics[width=0.25\linewidth]{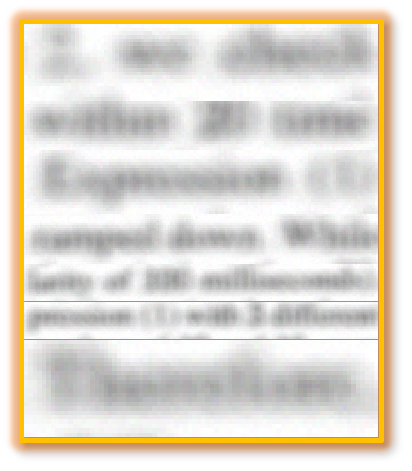}}
    \subfloat[\textbf{Ground-truth}]{\includegraphics[width=0.25\linewidth]{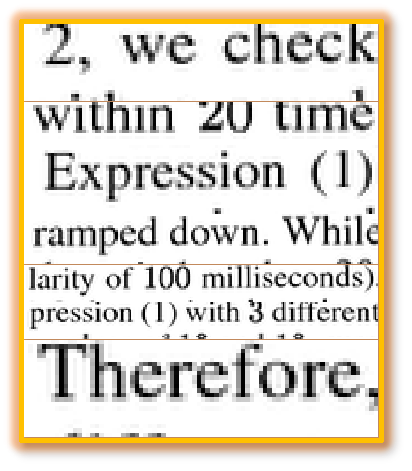}}
    \subfloat[\textbf{Coarse Predictor}]{\includegraphics[width=0.25\linewidth]{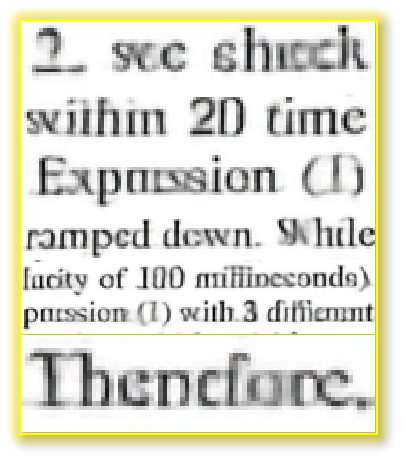}}
    \subfloat[\textbf{DocDiff-5}]{\includegraphics[width=0.25\linewidth]{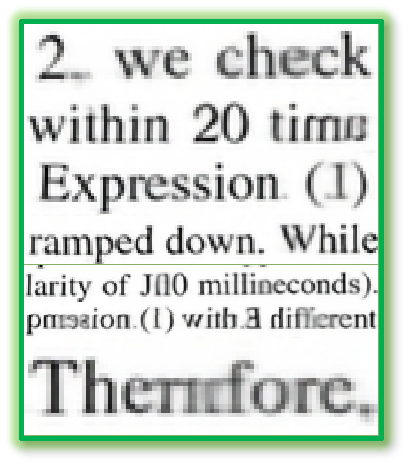}}
    \quad
    \subfloat[\textbf{DocDiff-10}]{\includegraphics[width=0.25\linewidth]{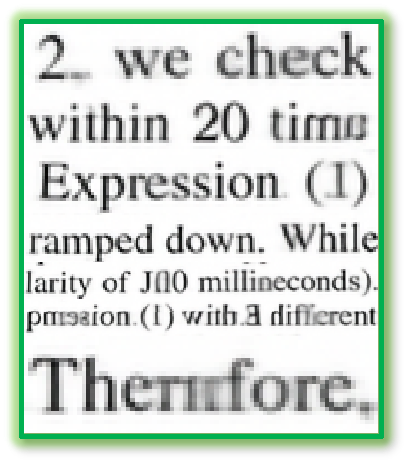}}
    \subfloat[\textbf{DocDiff-20}]{\includegraphics[width=0.25\linewidth]{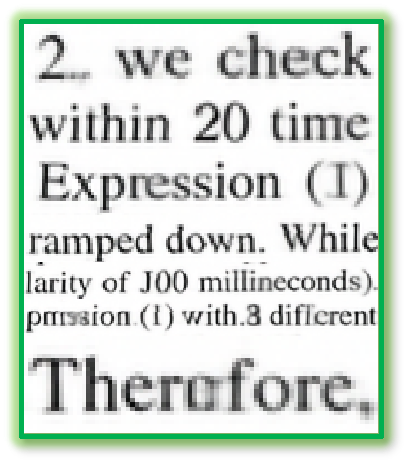}}
    \subfloat[\textbf{DocDiff-50}]{\includegraphics[width=0.25\linewidth]{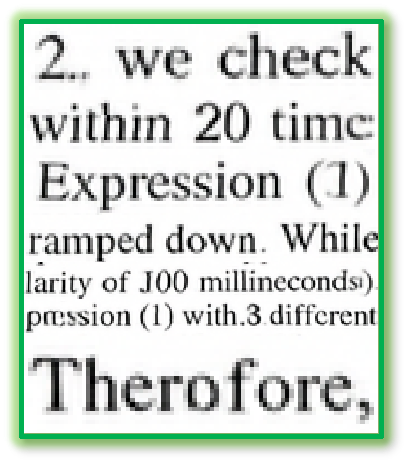}}
    \subfloat[\textbf{DocDiff-100}]{\includegraphics[width=0.25\linewidth]{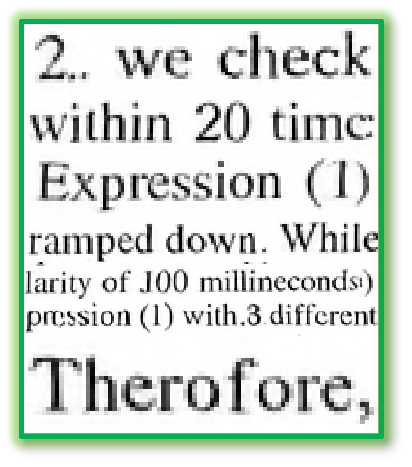}}
    \caption{Qualitative sampling results for different time steps on the Document Deblurring Dataset.}
    \label{fig:stage}
\end{figure}

Followed by DDIM, the time steps during training are set to 100, while different sampling steps are used during inference including 5, 10, 20, 50, and 100. Table \ref{tab:stage} shows quantitative results and Figure \ref{fig:stage} shows qualitative results. As the sampling step increases, there is an observable trend where the perceptual quality of the image improves, but this comes at the cost of increased distortion. As shown in Fig. \ref{fig:stage}, the edge of the word "Expression" become distinguishable after 5 sampling steps, while the edge of the word "Therefore" only become clear after 50 sampling steps. While DocDiff can correctly restore a majority of characters, errors can still occur such as "Therefore" becoming "Therofore". We introduce a possible solution to this problem in the next section. Even with 100 sampling steps during inference is usually considered a low sampling step for general diffusion models. We emphasize again that DocDiff is able to restore relatively sharp text edges within 20 steps thanks to its training strategy of predicting original data $x_0$ and its deterministic sampling strategy.

\begin{table}[!h]
\caption{Quantitative results for different sampling steps on the Document Deblurring Dataset. DocDiff-n means applying n-step sampling ($T$).}
\label{tab:stage}
\resizebox{\linewidth}{!}{%
\begin{tabular}{cccccccc}
\hline
 &  & \multicolumn{4}{c}{Perceptual} & \multicolumn{2}{c}{Distortion} \\ \cline{3-8} 
 &  & MANIQA↑ & MUSIQ↑ & DISTS↓ & LPIPS↓ & PSNR↑ & SSIM↑ \\ \hline
\multicolumn{2}{c}{DocDiff (Non-native)-5} & 0.6873 & 47.92 & 0.0907 & 0.0582 & \textbf{22.17} & \textbf{0.9223} \\
\multicolumn{2}{c}{DocDiff (Non-native)-10} & 0.6878 & 47.93 & 0.0886 & 0.0592 & 22.16 & 0.9217 \\
\multicolumn{2}{c}{DocDiff (Non-native)-20} & 0.6890 & 47.99 & 0.0875 & 0.0608 & 22.13 & 0.9206 \\
\multicolumn{2}{c}{DocDiff (Non-native)-50} & 0.6912 & 48.13 & 0.0776 & 0.0565 & 21.88 & 0.9180 \\
\multicolumn{2}{c}{DocDiff (Non-native)-100} & \textbf{0.6971} & \textbf{50.31} & \textbf{0.0636} & \textbf{0.0474} & 20.46 & 0.9006 \\ \hline
\end{tabular}%
}
\end{table}

\section{Future Works}

There are several potential solutions to overcome the limitations of our work. During the experiment, we observe that although DocDiff is able to generate sharp text edges in most cases, there are still instances where characters appear erroneous or distorted. To address this issue, one approach is to utilize the text prior in order to incorporate additional semantic information. Currently, there is a scarcity of paired large-scale document enhancement benchmark datasets in real-world scenarios. This leads to a lack of generalizability in practical applications. One approach to address this issue is to utilize DocDiff for assisted annotation. For instance, in the case of seal removal, humans can annotate on the results generated by DocDiff, thereby substantially reducing labelling complexity. Through multiple rounds of iteration and fine-tuning, the dataset can be expanded and the performance of the model can be improved.

\end{document}